\pdfoutput=1

\documentclass[11pt]{article}

\usepackage[]{emnlp2023}

\usepackage{times}
\usepackage{url}
\usepackage{latexsym}
\usepackage[font=small,labelfont=bf]{caption} 
\usepackage[inline]{enumitem}
\usepackage{booktabs}
\usepackage{lipsum,graphicx}
\usepackage{inconsolata}
\usepackage{microtype}
\usepackage{comment}
\usepackage{graphicx}
\usepackage{babel,blindtext,multicol,multirow}
\usepackage{geometry}
\usepackage{pdflscape}
\usepackage{rotating}

\usepackage{subfig}
\usepackage{subcaption}
\usepackage{adjustbox}
\usepackage{mathtools}
\usepackage{cleveref}

\DeclareMathOperator{\Tr}{Tr}

\let\oldbibliography\thebibliography
\renewcommand{\thebibliography}[1]{%
  \oldbibliography{#1}%
  \setlength{\itemsep}{0pt}%
}

\title{An Empirical study of Unsupervised Neural Machine Translation: analyzing NMT output, model's behavior and sentences' contribution}

\author{Isidora Chara Tourni \\
  Boston University\\
  \texttt{isidora@bu.edu} \\\And
  Derry Wijaya \\
  Boston University \\
  \texttt{wijaya@bu.edu} \\}

\date{}

\begin{document}
\maketitle
\begin{abstract}
\vspace{-2mm}
Unsupervised Neural Machine Translation (UNMT) focuses on improving NMT results under the assumption there is no human translated parallel data, yet little work has been done so far in highlighting its advantages compared to supervised methods and analyzing its output in aspects other than translation accuracy.
We focus on three very diverse languages, French, Gujarati, and Kazakh, and train bilingual NMT models, to and from English, with various levels of supervision, in high- and low- resource setups, measure quality of the NMT output and compare the generated sequences' word order and semantic similarity to source and reference sentences. We also use Layer-wise Relevance Propagation to 
evaluate the source and target sentences’ contribution to the result, expanding the findings of previous works to the UNMT paradigm.

\end{abstract}

\section{Introduction}
\vspace{-2.7mm}
Unsupervised Neural Machine Translation (UNMT) 
has been widely studied \citep{wang2021advances,marchisio2020does,kim2020and,lample2017unsupervised,artetxe2019effective,su2019unsupervised}, in an effort to create efficient and trustworthy NMT models of excellent performance not relying on the existence of parallel data. Obtaining high-quality parallel corpora is expensive and time-consuming, especially for less-common language pairs. Unsupervised NMT hence aims to circumvent this limitation.
NMT in general significantly aids in preserving indigenous languages by making global information accessible, supports migrants in overcoming language barriers to essential services, and enables the globalization of local news from smaller countries. 
UNMT particularly has broad applicability, especially in addressing linguistic diversity and information accessibility challenges.

However, there has been little effort on analyzing, apart from the quality of the output, the model behavior during UNMT, and 
the models' inner workings and the effects of various setups on hypotheses and generated translations' quality, monotonicity and semantic similarity, as well as model robustness and consistency.
We analyze and compare 
UNMT approaches for two very diverse languages, French and Gujarati, translating to and from English. 
We research into the existence of different stages in UNMT, analyze source and target sentence tokens' contributions 
to the result 
\citep{bach2015pixel} 
evaluate the quality and word alignment of generated translations, and Robustness and Consistency of our model to perturbed inputs. Our paper follows up closely on the work of \citet{voita2020analyzing,voita2021language,marchisio2022systematic}, and examines the following questions:
\begin{itemize}
\item Do the distinct stages of transformer-based NMT analyzed in previous works exist in Unsupervised, and joint Supervised \& Unsupervised NMT? 
\item How does output quality, word alignment, 
semantic similarity, 
as well as source and target sentences' token contributions to the NMT output behave across the aforementioned stages?
\item How Robust and Consistent are NMT models throughout training?
\end{itemize}

Our findings confirm the existence of NMT stages regardless of the level of training supervision, and show that Unsupervised methods 
produce translations more similar to source sentences in terms of word order, 
yet more semantically distant. UNMT models tend to show higher Robustness and Consistency, and can more easily recover from sentence perturbations. 
We also observe that in reduced training data experiments, there is a heavy reliance on the source sentence for generating translations. Our focus is not on outperforming NMT state-of-the-art results, but rather on training and examining the behavior of bilingual models in various setups. 

The paper is structured as follows: in Section \ref{related} we present related work in the topics of UNMT, NMT analysis and other metrics analyzed in our work. In Section \ref{methods_experiments} we analyze the methods proposed for NMT analysis and the experiments conducted, while in Section \ref{results_discussion} we present and discuss our findings. Finally, in Sections \ref{conclusions}, \ref{limitations} and \ref{ethics} we conclude our work and highlight certain limitations and ethical considerations, respectively. We present additional experiments and results on the Robustness of the models and Semantic Similarity of input and output sentences in the appendix.

\section{Related Work}
\label{related}
\vspace{-2.7mm}
\subsection*{Unsupervised Neural Machine Translation}
\vspace{-1mm}
UNMT aims to make NMT work in the absence of parallel data. Most common approaches have focused on cross- or multilingual initialization of a language model either through an alignment of monolingual embeddings \citep{artetxe2017unsupervised,lample2018phrase,conneau2017word,lample2017unsupervised} or by model pretraining and fine-tuning
\citep{lample2019cross,song2019mass,liu2020multilingual}. Back-Translation (BT) \citep{sennrich2015improving} 
translates monolingual data between languages, 
creating pseudo-parallel training corpora \citep{artetxe2017unsupervised,lample2017unsupervised}.
\newcite{marchisio2022systematic} first systematically examine the naturalness and diversity of the UNMT output, comparing it to similar quality human translations, and proposing a way to leverage UNMT to improve a classical supervised NMT system.
In more recent works,
\citet{liu2022flow} introduce a flow-adapter architecture 
to separately model the distributions of source and target languages, 
and \citet{he2022bridging} identify and mitigate a training and inference style and content gap between back-translated data and natural source sentences. 
\newcite{garcia2020multilingual} expand the paradigm to multilingual UNMT, 
while \citet{garcia2020harnessing} use offline BT synthetic data 
to improve multilingual En-xx UNMT for low-resource languages xx. 

\subsection*{Layer-wise Relevance Propagation (LRP)}
\vspace{-1mm}
LRP \citep{bach2015pixel} measures relevance of the input components, or the neurons of a network, to the next layers' output, and is directly applicable to layer-wise architectures.
\newcite{wu2021explaining} use LRP as an attribution method 
for sequence classification tasks. 
We extend its usage to the Transformer, 
and measure the relevance of source and target sentences to the NMT output.

\subsection*{Neural Machine Translation analysis}
\vspace{-1mm}
\newcite{voita2020analyzing} examine the source and target sentences' tokens' relative contributions to NMT output, adapting LRP to a Transformer, and experimenting with different training objectives, training data amounts and types of target sentence prefixes, and their effect on NMT output quality and monotonicity. Following up to that work,
\citet{voita2021language} analyze NMT stages, drawing parallels to distinct SMT stages.  
Their findings include decomposing NMT into three phases, 
and using the key learning advantages of each stage to improve non-autoregressive NMT.  We examine and identify if those stages exist in UNMT. 

\subsection*{Robustness \& Consistency}
\vspace{-1mm}
Previous works examine Robustness in NLP \citep{yu2022measuring,wang2021measure,la2022king}, measuring and improving NLP models' performance against perturbed or unseen input. 
Specifically for NMT, 
\newcite{niu2020evaluating} propose two metrics, Robustness and Consistency to measure sensitivity of a model to input perturbations. 

\section{Method \& Experiments}
\label{methods_experiments}
\subsection{Model}
\vspace{-1mm}
We use a 6-layers 8-heads transformer-based model, XLM \citep{lample2019cross}, following the training configurations and hyperparameters suggested by the authors. We use Byte Pair Encoding to extract a 60k vocabulary, an embedding layer size of 1024, a dropout value and an attention layer dropout value of 0.1, and a sequence length of 256. We measure the quality of the Language Model (LM) with perplexity, and quality of the NMT output with BLEU 
both used as training stopping criteria, when there is no improvement over 10 epochs. 
We first pre-train a LM in each language with the MLM objective, and use it to initialize the encoder and decoder of the NMT model. We then train NMT models, 
using Back-Translation (BT) and denoising auto-encoding (AE) with the monolingual data used for LM pretraining for UNMT, the Machine Translation (MT) objective for the Supervised NMT model, and BT-MT for the joint Unsupervised and Supervised approach.

\vspace{-2mm}
\subsection{Datasets}
\vspace{-1mm}
The languages we work with are English, French, Gujarati, and Kazakh 
and we're translating in all directions, English--French (En--Fr), French--English (Fr--En), English--Gujarati (En--Gu), Gujarati--English (Gu--En), English--Kazakh (En--Kk), Kazakh--English (Kk--En). 
For English and French, we use 5 million News Crawl 2007-2008 monolingual sentences for each language, and 23 million WMT14 parallel sentences. 
For Gujarati, we have 1.4 million sentences and for Kazakh we have 9.5M monolingual sentences, collected for both languages from Wikipedia, WMT 2018, 2019 and Leipzig Corpora (2016)\footnote{https://wortschatz.unileipzig.de/en/download/}. 
As parallel data, we have 22k sentences 
from the WMT 2019 News Translation Task\footnote{http://data.statmt.org/news-crawl/} for Gu--En and Kk--En, respectively/ 
As development and test sets, we use newstest2013 and newstest2014, respectively, for En--Fr and Fr--En, WMT19 for En--Gu and Gu--En and En--Kk and Kk--En. 


\vspace{-2mm}
\subsection{Layer-wise Relevance Propagation}
\vspace{-1mm}
\newcite{voita2020analyzing} explain how LRP calculation in a Transformer is confusing due to the non-clear layered nature of the model. We follow their setup, with LRP to be propagated first inversely through the decoder and the encoder, up to the input model layer, and without assuming the conservation principle holds per layer, but only across processed tokens. LRP is the relevance of input neurons to the top-1 logit predicted by the model, and token contribution is the sum of the input neurons' relevance.
Total source and target sentence contributions to the result at generation step t are given by 
$R_t(source)  = \sum_{i}^{}{R(x_i)}$, 
$R_t(target) =  \sum_{j=1}^{t-1}{R(y_j)}.$
At every step t, Relevance follows the conservation principle:
$R_t(source) + R_t(target) = 1$. At step 1, we have $R_1(source) = 1$, 
$R_1(target) = 0$. For every target token past the currently generated one, LRP is 0.

\vspace{-2mm}
\subsection{Word Order}
\vspace{-1mm}
Our aim is to examine differences in word order between translations and reference or source sentences. We evaluate two different reordering metrics, Fuzzy Reordering Score (FRS) \citep{talbot2011lightweight,nakagawa2015efficient}\footnote{https://github.com/google/topdown-btg-preordering} and Translation Edit Rate (TER) \citep{snover2006study}.
FRS ranges between 0 and 1, with larger values for highly monotonic alignments (higher structural similarity and closer word order). For a translation ${y'}$ and reference ${y}$ (or source sentence ${y}$):
$FRS(y',y) = 1 - \frac{C-1}{M-1}.$
C is the number of chunks of contiguously
aligned translation words, intuitively perceived as the number of times a reader would need to jump in order to read the system's reordering of the sentence in the order proposed by the reference of length M.
With \textit{fast\_align}\footnote{https://github.com/clab/fast\_align} we calculate word alignments.
TER is defined as
$TER(y',y) = \frac{E}{L_y}, $
where E is the number of edits needed to modify the produced translation $y'$ to match the reference sentence $y$ (or source sentence $y$), and $L_{y}$ is the length of the reference (or the source sentence, respectively). TER ranges between 0 and 1, and low values indicate more monotonic alignments.

\subsection{Model Robustness}
\vspace{-1mm}
\newcite{niu2020evaluating} set \textit{TQ} (y', y) to be the model quality of model M with translation $y'$, and reference $y$, to define the concepts of Consistency and Robustness.
Consistency of a model M on input $x$ and perturbation $\delta$ is given by 
$CONSIS(M|x,\delta) = \textit{Sim}(y\textsubscript{$\delta$},y), $
where $y$ is the reference translation, and y\textsubscript{$\delta$} is the translation of the perturbed sentence. \textit{Sim} is the harmonic mean between model quality \textit{TQ}(y', y\textsubscript{$\delta$}') and \textit{TQ}(y\textsubscript{$\delta$}', y'), measured between translations y', y\textsubscript{$\delta$}', respectively.
On the other hand, Robustness of a model M is defined as the ratio between quality of the model producing translations y\textsubscript{$\delta$}' and y':
$ROBUST(M|x,\delta) = \frac{TQ(y_{\delta}',y)}{TQ(y',  y)}.$
It takes values in [0,1].
We evaluate Consistency and Robustness on test sets perturbed with two different approaches: a. \textbf{misspelling} - each word is misspelled (random deletion, insertion or substitution of characters) with a probability of 0.1, b. \textbf{case-changing} - each sentence is modified (upper-casing, lower-casing or title-casing all letters) with a probability of 0.5.

\subsection{Semantic Similarity}
\vspace{-1mm}
In evaluating semantic similarity between human translations or source sentences and generated translations, we calculate the Ratio Margin-based Similarity Score (RMSS) between each reference or source sentence and its k-nearest neighbors among all translations 
\citep{artetxe2018margin}.
We follow \citet{keung2020unsupervised} to obtain and mean-pool the
mBERT 
embedding vectors of all 
sentences. 
We set $cos(\cdot,\cdot)$ to be the cosine similarity and NN$_{k}^{src}(x)$ the $k$ nearest neighbors of $x$ in the reference or source sentence embedding space. RMSS is high when source and target pairs compared are closer than their respective nearest neighbors. For a hypothesis $y$: 
\vspace{-2mm}
\begin{equation*}
\begin{multlined}
RMSS (x,y) = \\
\frac{cos(x,y)}{\sum_{z \in \text{NN}^{tgt}_{k}(x)} \frac{cos(x,z)}{2k} + \sum_{z \in \text{NN}^{src}_{k}(y)} \frac{cos(y,z)}{2k}}.
 \end{multlined}
\end{equation*}

\section{Results \& Discussion}
\label{results_discussion}
\begin{table*}[ht]
\vspace{-1mm}
\small
\captionsetup{font=scriptsize}
\centering
\begin{tabular}{lrrrrrrr}
\toprule
\textbf{Method}  & \textbf{en-fr}  & \textbf{fr-en} &\textbf{en-gu} & \textbf{gu-en} &\textbf{en-kk} & \textbf{kk-en} \\ 
\midrule
\multicolumn{7}{l}{\emph{Other methods}} \\ 
\midrule
 &45.9\footnote{https://www.deepl.com/press.html}   &-  &0.1 \footnote{https://github.com/google-research/bert/blob/master/multilingual.md}
 &0.3 \footnote{https://github.com/google-research/bert/blob/master/multilingual.md}&2.5 \footnote{https://github.com/google-research/bert/blob/master/multilingual.md} &7.4 \footnote{https://github.com/google-research/bert/blob/master/multilingual.md}&  \\ 
 \midrule
 \midrule
\multicolumn{7}{l}{Our method} \\ \hline
BT+AE    &21.76   &21.69&0.4 &0.46&0.7&1.0 \\ 
\midrule
\midrule
Parallel data: 22k&&&&&&&\\
\midrule
MT    & 31.12 &30.63 & 1.04 &\textbf{2.65}&2.4&2.6& \\ 
BT+AE+MT  & 34.54  & 34.02& \textbf{1.16} & 2.19&2.8&2.9& \\ 
\midrule
\midrule
132k&&&&&&&\\
\midrule
MT &37.8&35.6&-&-&5.2&8.0& \\ 
BT+AE+MT &38.6&38.4&-&-&\textbf{6.6}&\textbf{8.9}& \\
\midrule
\midrule
1m&&&&&&&\\
\midrule
MT   &41.25&41.33& - & - &-&- \\
BT+AE+MT   &40.37&40.4& - & -&-&- \\
\midrule
\midrule
2.5m&&&&&&&\\
\midrule
MT   &40.46&40.71& - & - &-&- \\
BT+AE+MT   &39.88&39.62& - & - &-&-\\
\midrule
\midrule
5m&&&&&&\\
\midrule
MT   & 41.52  & 41.18& - & - &-&- \\ 
BT+AE+MT   & 40.89  & 40.8& - & -&-&-\\
\midrule
\midrule
23m&&&&&&\\
\midrule
MT   &\textbf{41.75}&\textbf{41.41}& - & -&-&-  \\
BT+AE+MT &41.29&40.99& - & -&-&- \\
\bottomrule
\end{tabular}
\vspace*{0.5em}
\caption{\scriptsize{BLEU scores for En--Fr, Fr--En, En--Gu, Gu--En, En--Kk, Kk--En NMT. 
Test and validation sets are WMT19 for Gujarati and Kazakh, newstest2013-14 for French pairs. State-of-the art results given for the sake of consistency. \emph{MT} stands for machine translation objective, \emph{BT} stands for Back-Translation and \emph{AE} for denoising auto-encoding.}}
\vspace{-2mm}
\label{table:bleu_results}
\vspace{-2mm}
\end{table*}

BLEU scores of converged models are seen in Table \ref{table:bleu_results}. 
In En--Gu, En--Kk, Kk--En, BT-MT improves BLEU; 
Models trained with parallel data show higher BLEU, and 
absence of parallel data or often introduction of low quality data (eg in Gu--En) through BT lowers output quality.

\begin{figure*}[t!]
\centering
\captionsetup{font=scriptsize}
\begin{multicols}{3}
    \centering
    \includegraphics[width=.33\textwidth]{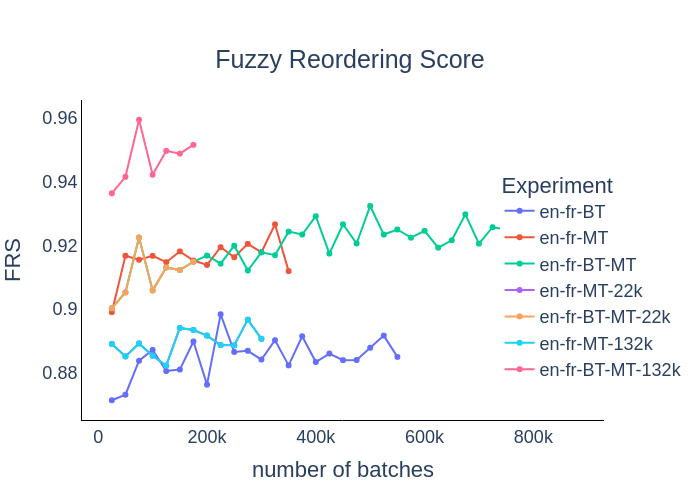}
    \includegraphics[width=.33\textwidth]{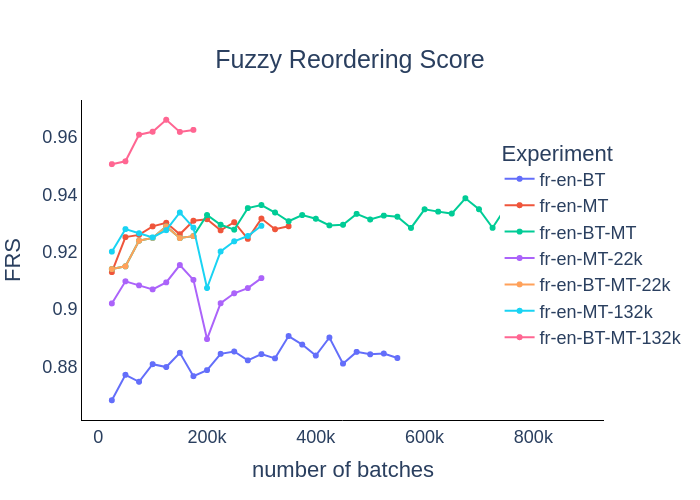}
    \includegraphics[width=.33\textwidth]{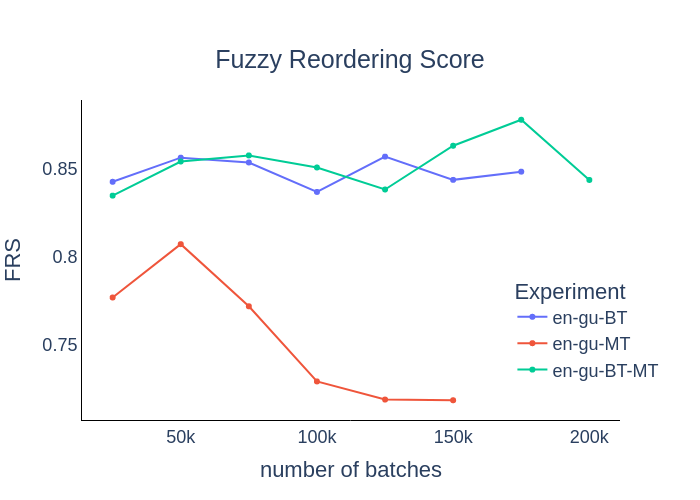}
    \includegraphics[width=.33\textwidth]{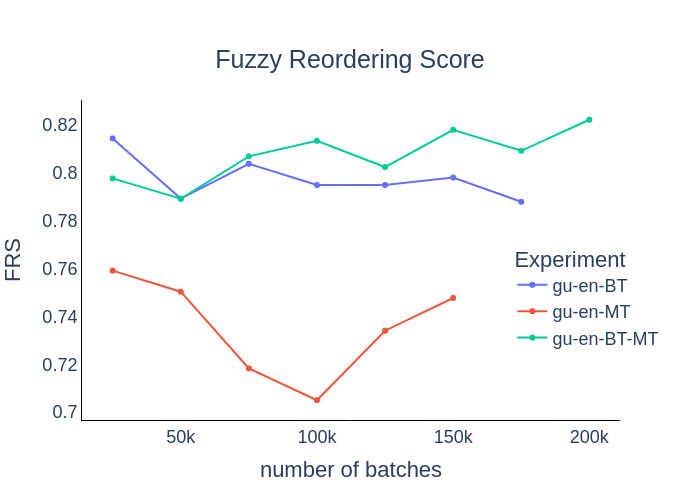}
    \includegraphics[width=.33\textwidth]{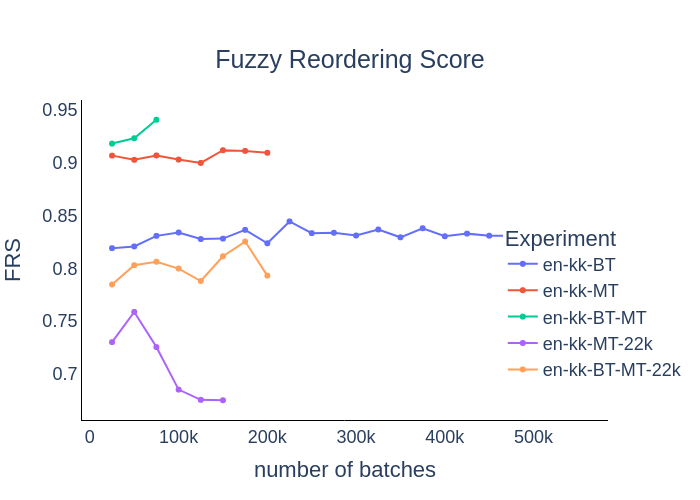}
    \includegraphics[width=.33\textwidth]{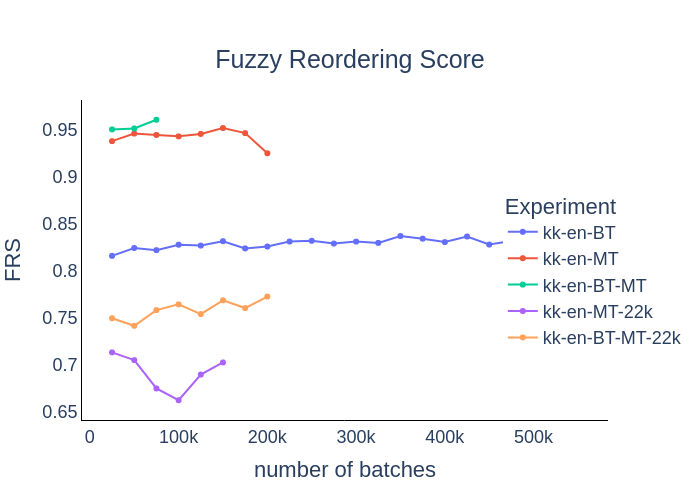}
\end{multicols}
\vspace{-6mm}
\caption{Fuzzy Reordering Scores (FRS) between references and generated translations, for En--Fr, Fr--En, En--Gu, Gu--En, En--Kk, Kk-En during training.}
\label{fig:frs_over_batches}
\vspace{-4mm}
\begin{multicols}{3}
    \centering
    \includegraphics[width=.33\textwidth]{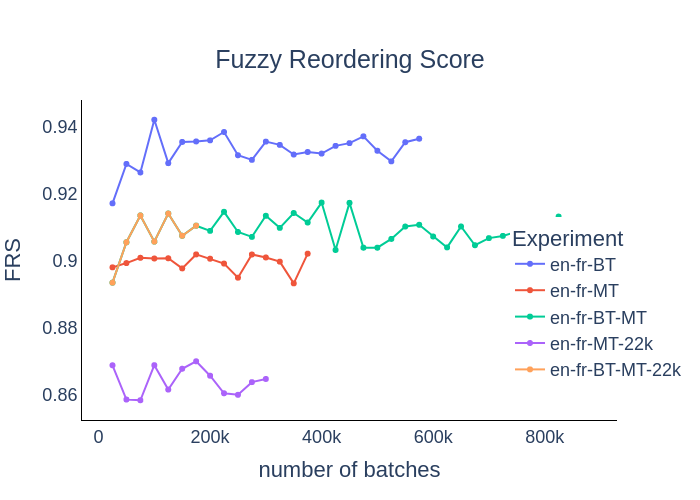}
    \includegraphics[width=.33\textwidth]{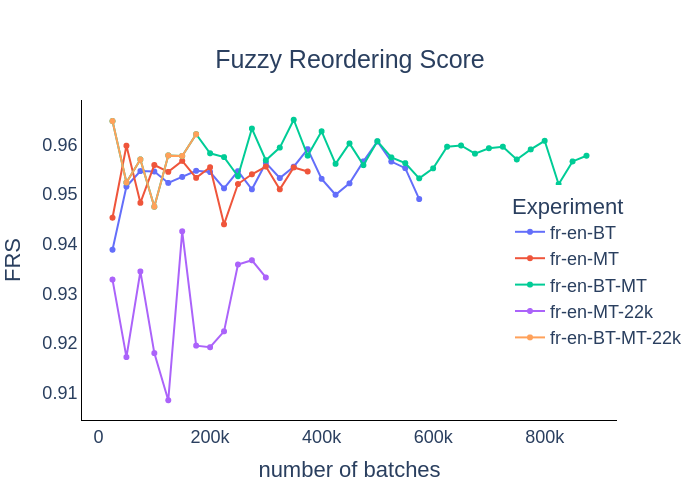}
    \includegraphics[width=.33\textwidth]{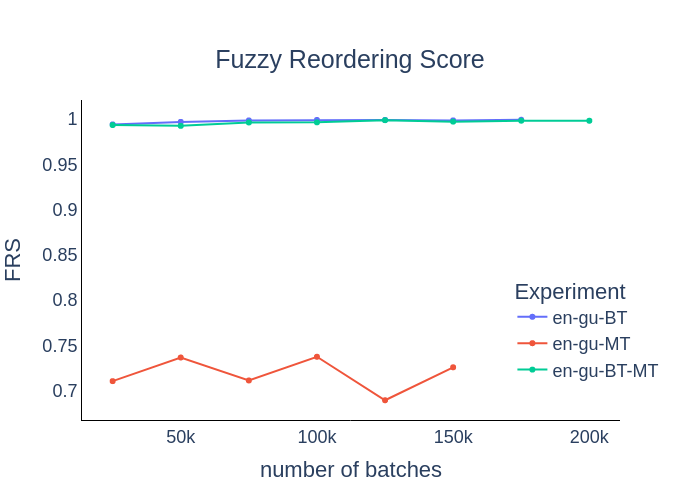}
    \includegraphics[width=.33\textwidth]{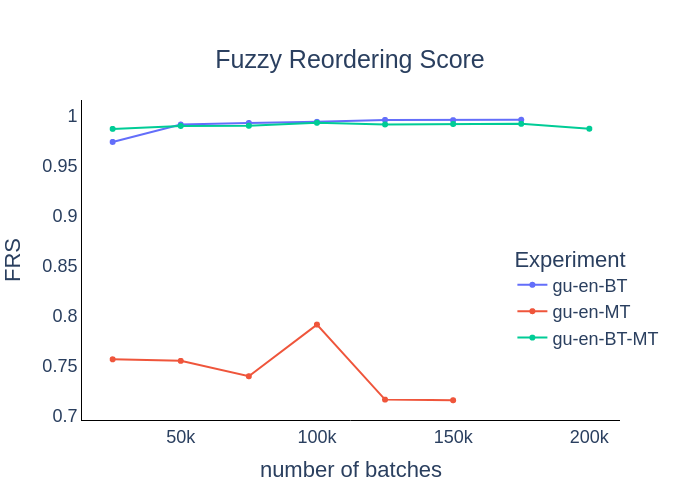}
    \includegraphics[width=.33\textwidth]{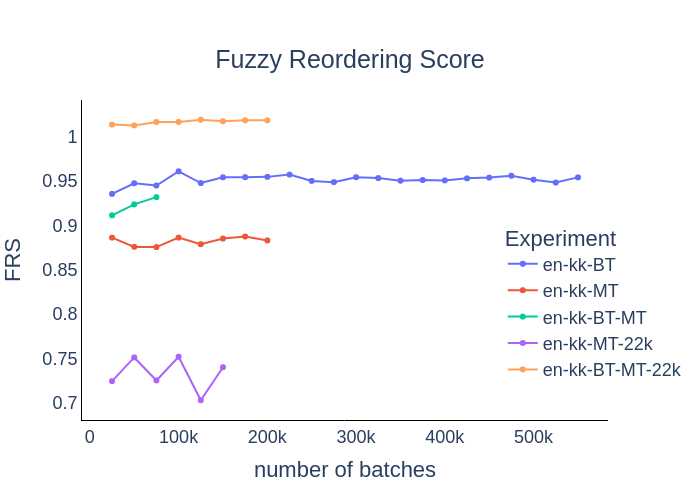}
    \includegraphics[width=.33\textwidth]{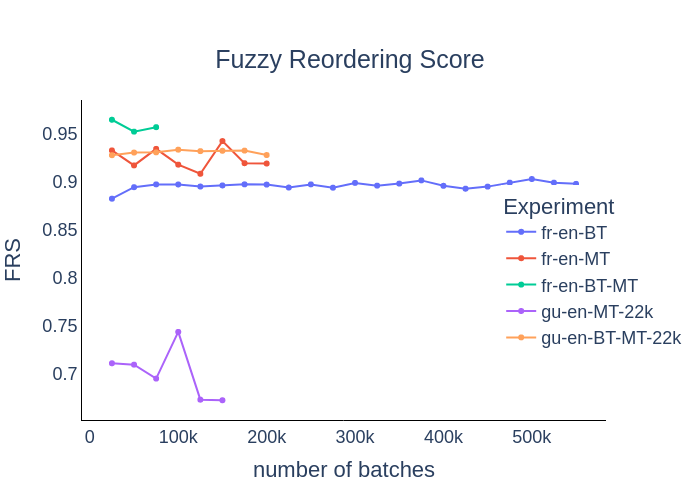}
\end{multicols}
\vspace{-6mm}
\caption{Fuzzy Reordering Scores (FRS) between source sentences and generated translations for En--Fr, Fr--En, En--Gu, Gu--En, En--Kk, Kk--En during training.}
\label{fig:frs_diff_over_batches}
\end{figure*}

\subsection*{FRS}
\vspace{-1mm}
In most En--Fr, Fr--En experiments, there is a large fluctuation yet a small and gradual FRS increase between translations and references (Fig. \ref{fig:frs_over_batches}), and then a small decrease. 
Higher FRS shows more monotonic alignments. Starting from non-monotonic alignments in the first stage, we get maximum FRS values in the second stage of training - highly aligned translations and references - which slightly decrease in the third stage until model convergence.
With parallel data we get the most monotonic alignments, while we have the least identical reorderings between references and translations in BT-only cases, in both high- and low-resource setups. Similar patterns are observed in En--Kk and Kk--En, where we have the least monotonic alignments for few/no parallel data (MT-22k, BT-MT-22k, BT), and the most for experiments with parallel data (MT, BT-MT), with values slightly increasing and then remaining stable throughout training in most cases.
BT En--Gu, Gu--En 
models show high and steady FRS values: between languages with a complicated and non-monotonic alignment, BT produces translations more aligned with the reference. 

For En--Fr, Fr--En, FRS (Fig. \ref{fig:frs_diff_over_batches}) values are stable throughout training, and BT, BT-MT experiments' results imply highly monotonic alignments; with BT, translations are closer to source sentences in terms of word order. FRS is lower in MT only experiments, as source and translation alignments are less monotonic when models are trained with parallel data alone.
Results are similar in  En--Gu, Gu--En. 
BT, BT-MT give an 
almost perfect alignment between source sentences and translations.

For En--Kk, Kk--En, we observe that in the majority of experiments, source sentences are highly monotonic to translations, with steady FRS values throughout training.

We see that BT yields more stable and higher alignment scores compared to models trained only on parallel data, suggesting it offers a significant advantage for improving translation quality.

\begin{figure*}[t!]
\vspace{-2mm}
\captionsetup{font=scriptsize}
\begin{multicols}{3}
    \includegraphics[width=.33\textwidth]{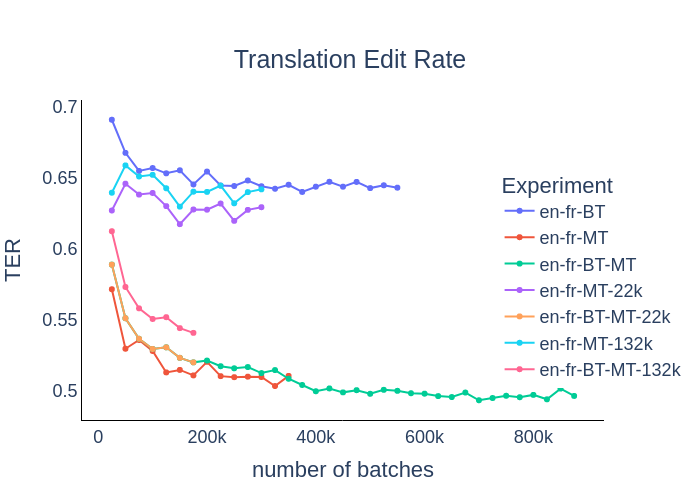}
    \includegraphics[width=.33\textwidth]{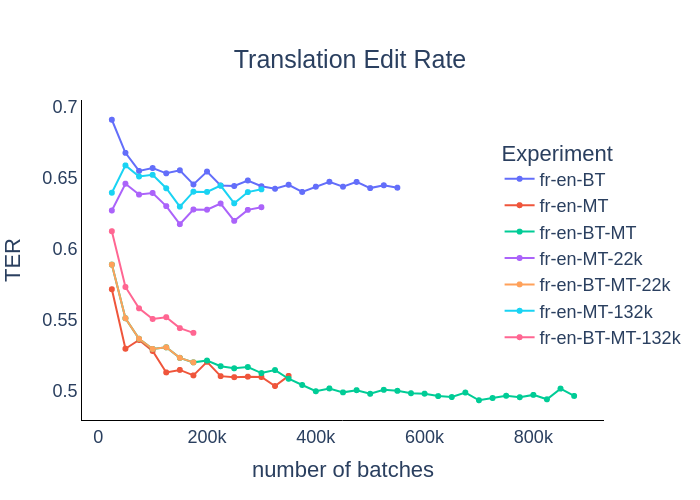}\\
    \includegraphics[width=.33\textwidth]{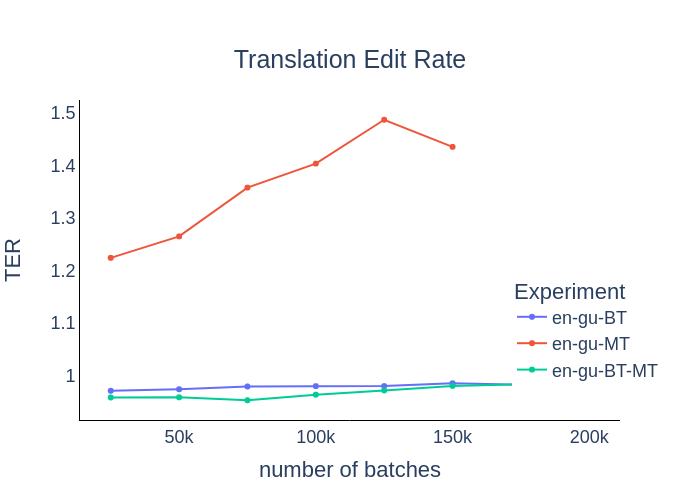}
    \includegraphics[width=.33\textwidth]{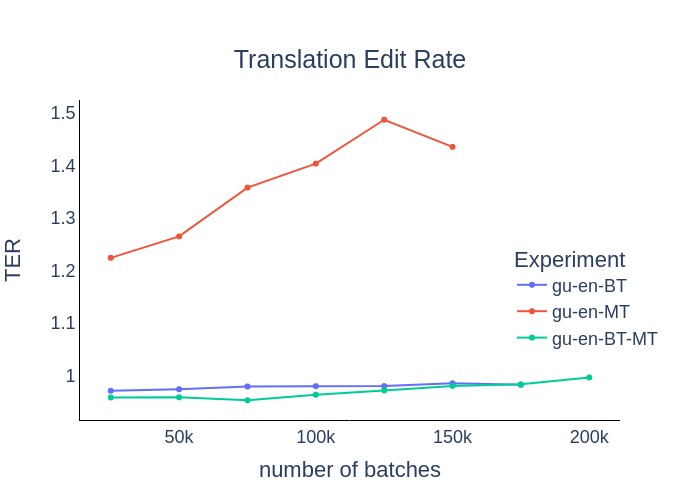}
    \includegraphics[width=.33\textwidth]{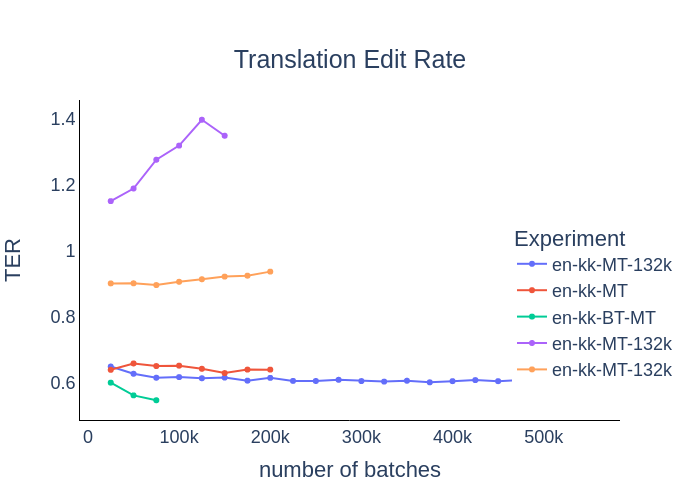}
    \includegraphics[width=.33\textwidth]{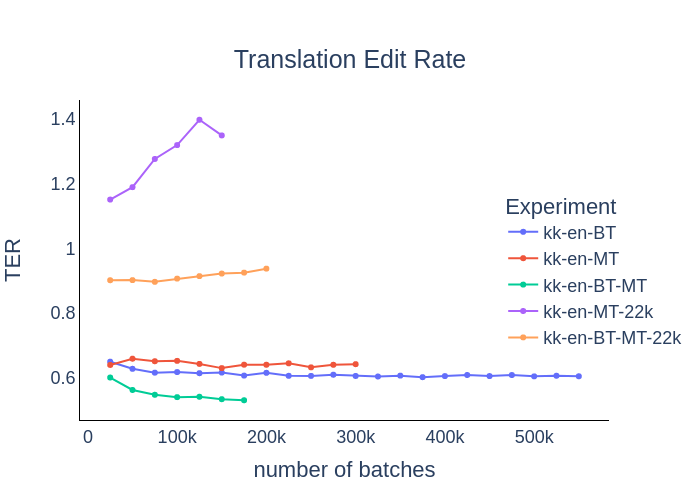}
\end{multicols}
\vspace{-6mm}
\caption{\label{fig:ter_over_batches}Translation Edit Rate (TER) between references and generated translations for En--Fr, Fr--En, En--Gu, Gu--En, En--Kk, Kk--En during training.}
\vspace{-4mm}
\begin{multicols}{3}
    \includegraphics[width=.33\textwidth]{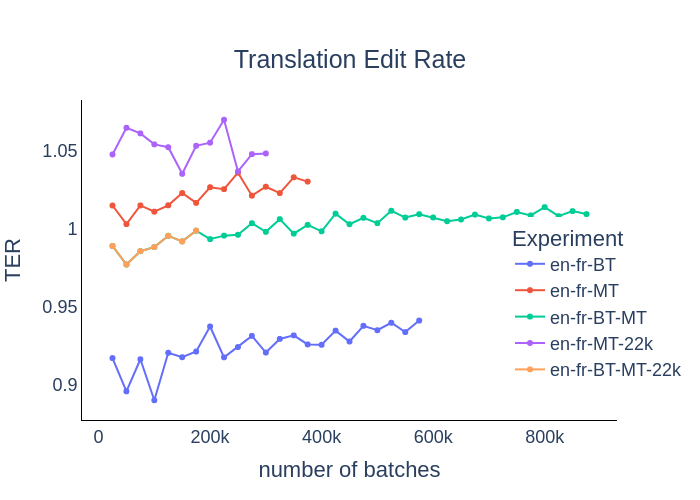}
    \includegraphics[width=.33\textwidth]{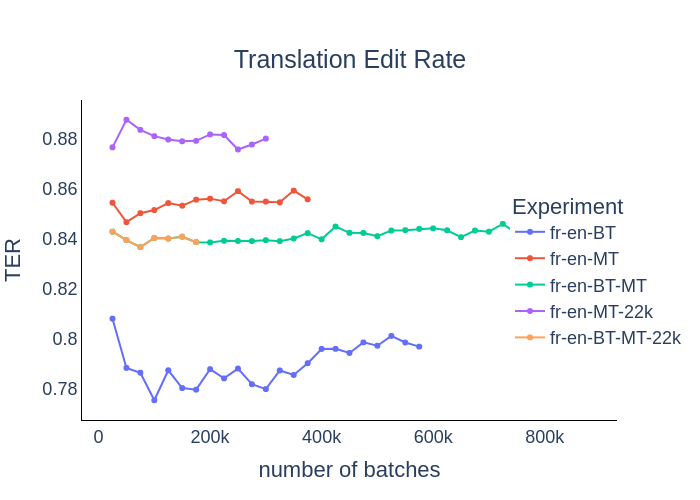}
    \includegraphics[width=.33\textwidth]{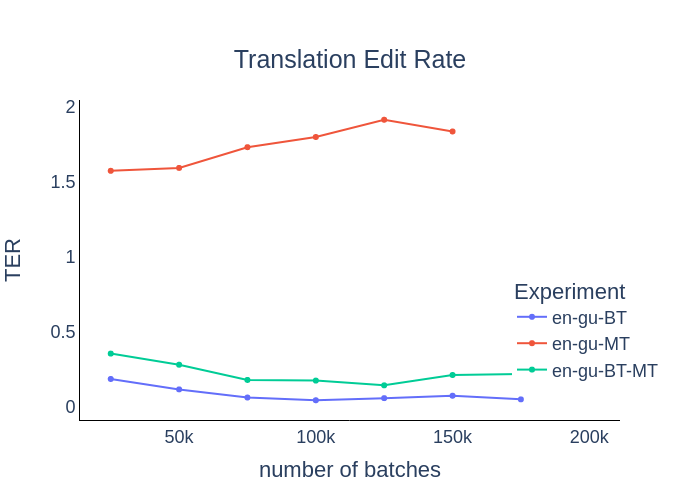}
    \includegraphics[width=.33\textwidth]{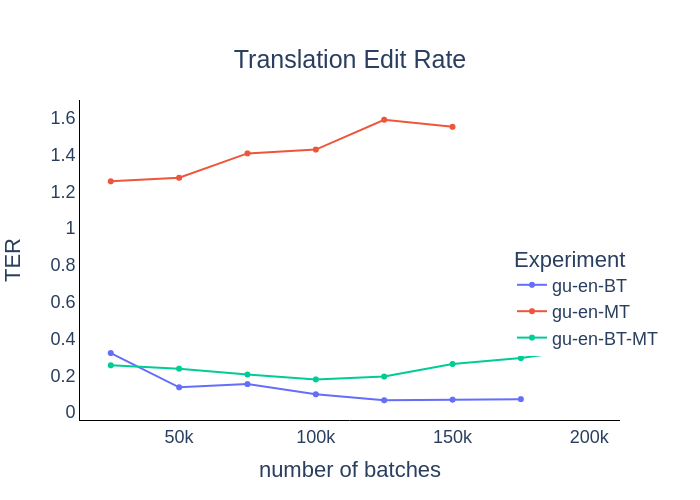}
    \includegraphics[width=.33\textwidth]{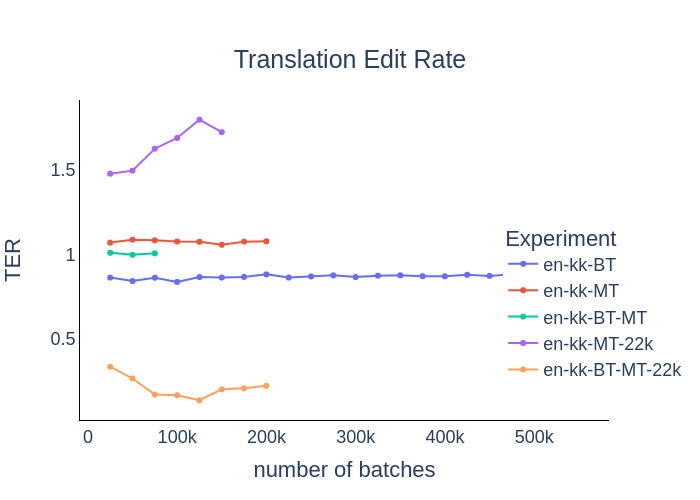}
    \includegraphics[width=.33\textwidth]{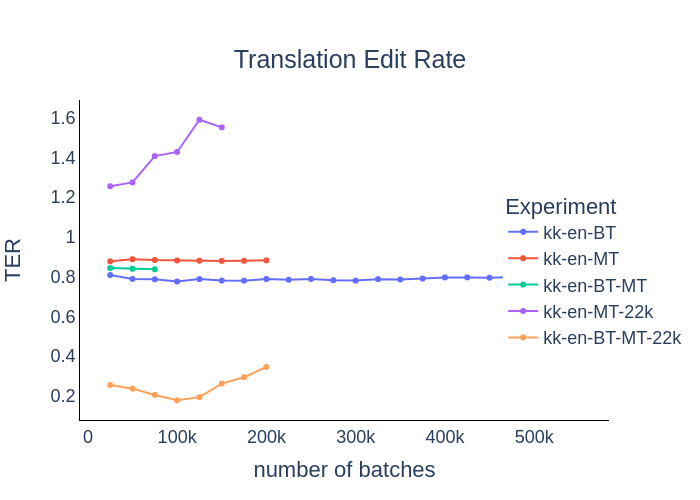}
\end{multicols}
\vspace{-6mm}
\caption{\label{fig:ter_diff_over_batches}Translation Edit Rate (TER) between source sentences and generated translations for En--Fr, Fr--En, En--Gu, Gu--En, En--Kk, Kk--En during training.}
\end{figure*}

\subsection*{TER}
\vspace{-1mm}
Observing TER between translations and references (Fig. \ref{fig:ter_over_batches}), 
in En--Fr and Fr--En, low TER for MT, BT-MT means more monotonic alignments, in contrast to higher TER in low-resource and BT-only experiments.
TER gradually decreases 
for all models, 
as sentences generated at the end of training highly resemble human translations.
Results are different for En--Gu and Gu--En; TER is low in BT- only and BT-MT models, but rather high, and increasing, in the MT-only model, with translations in the former case very close to references. BT produces more monotonic to the reference translations for a language diverse in terms of script, morphological complexity and word order from English.

Stable or slightly increasing TER values between source sentences and translations (Fig. \ref{fig:ter_diff_over_batches}) mean high structural resemblance. 
For En--Fr and Fr--En, BT and BT-MT show the lowest TER values, hence generated with BT sequences and source sentences have high monotonicity. Similarly, in En--Gu and Gu--En, BT, BT-MT models show lower TER and higher and more monotonic alignment of translations and source sentences.

For En--Kk, Kk--En, we see that we have higher monotonicity in higher/no supervision experiments, and lower in low-resource models- we can assume training with few parallel data and for few epochs highly cannot help the model properly align produced translations and references or source sentences. Between translations and source sentences, when the model is sufficiently trained, we surprisingly observe high monotonicity across experimental setups. 

We deduce that for En--Fr, Fr--En, translations have higher monotonicity to references in MT, BT-MT, lower in BT-only experiments, but higher to source sentences in BT, BT-MT and lower in MT. Training supervision leads to better translation to hypothesis alignment, while BT induces better translation to source sentence alignment. 

Hence, we see that the effectiveness of different training methods like MT and BT varies by language pair, with BT showing particular promise for languages are structurally diverse from English.


\subsection*{LRP analysis}
\begin{figure*}[t!]
\captionsetup{font=scriptsize}
\begin{multicols}{3}
    \includegraphics[width=.33\textwidth]{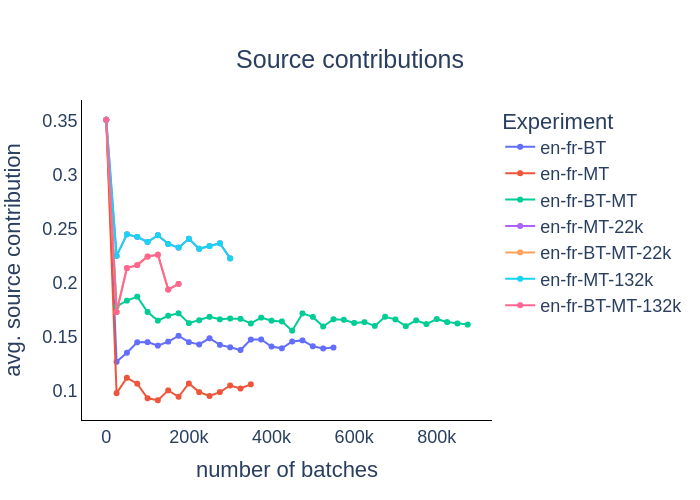}
    \includegraphics[width=.33\textwidth]{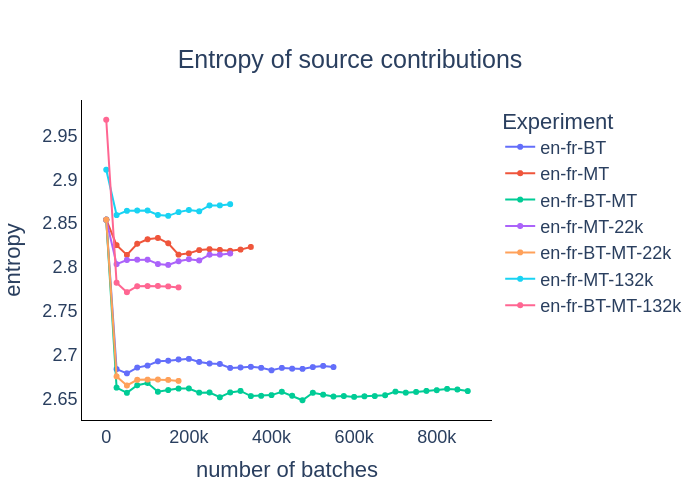}
    \includegraphics[width=.33\textwidth]{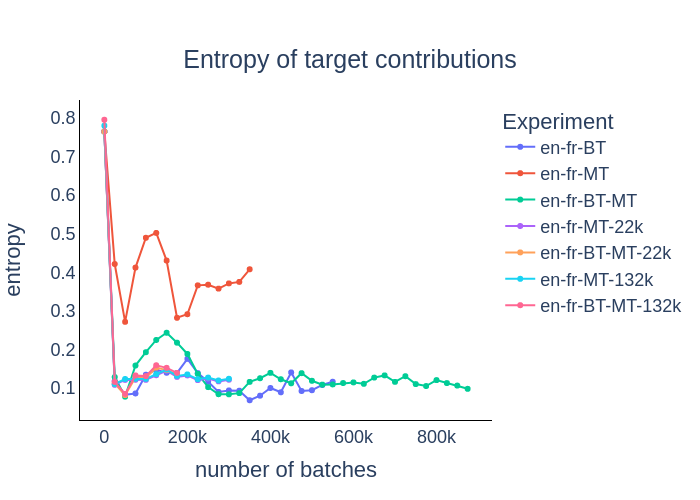}
\end{multicols}
\vspace{-7mm}
\caption{\label{fig:contrib1} En--Fr Average Source Contribution, Entropy of Source Contributions and Entropy of Target Contributions during training.}
\vspace{-2mm}
\begin{multicols}{3}
    \includegraphics[width=.33\textwidth]{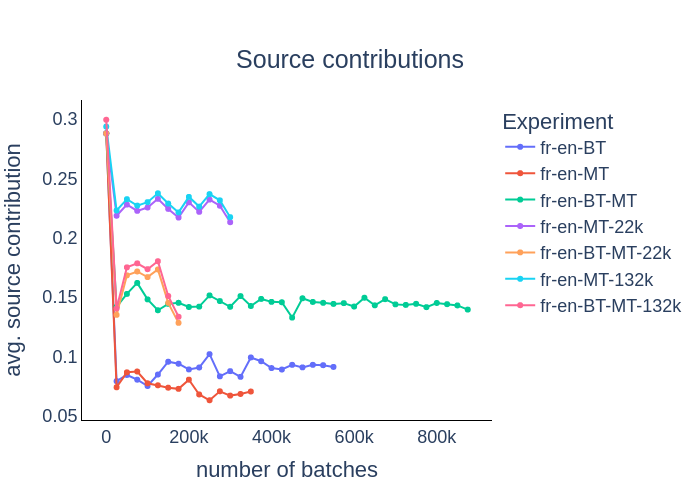}
    \includegraphics[width=.33\textwidth]{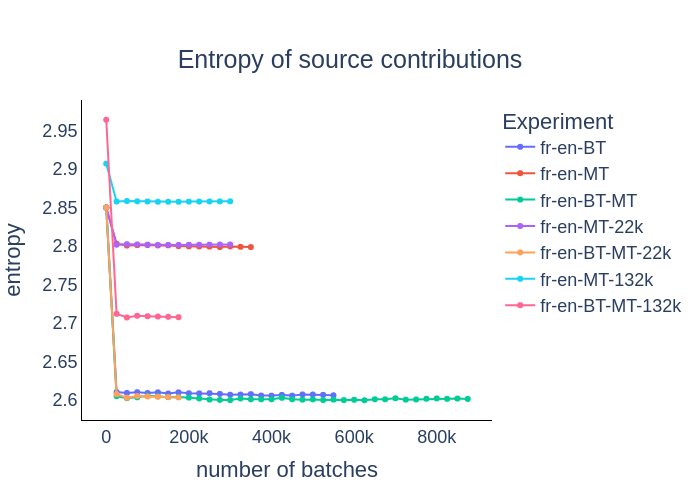}
    \includegraphics[width=.33\textwidth]{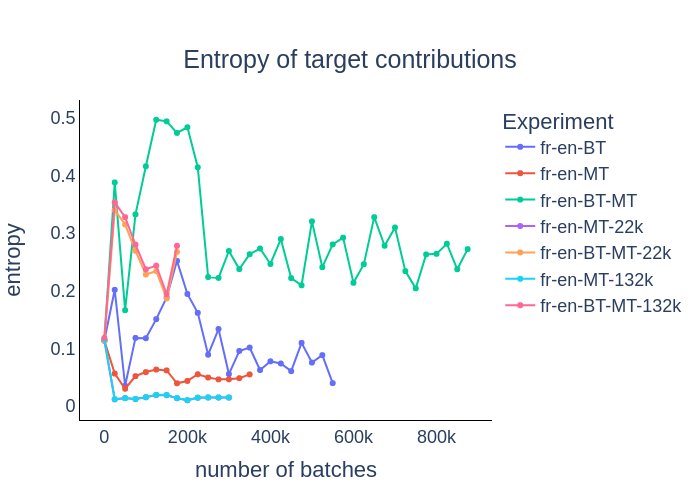}
\end{multicols}
\vspace{-7mm}
\caption{\label{fig:contrib2}Fr--En Average Source contribution, Entropy of Source Contributions and Entropy of Target Contributions during training.}
\vspace{-4mm}
\begin{multicols}{3}
    \includegraphics[width=.33\textwidth]{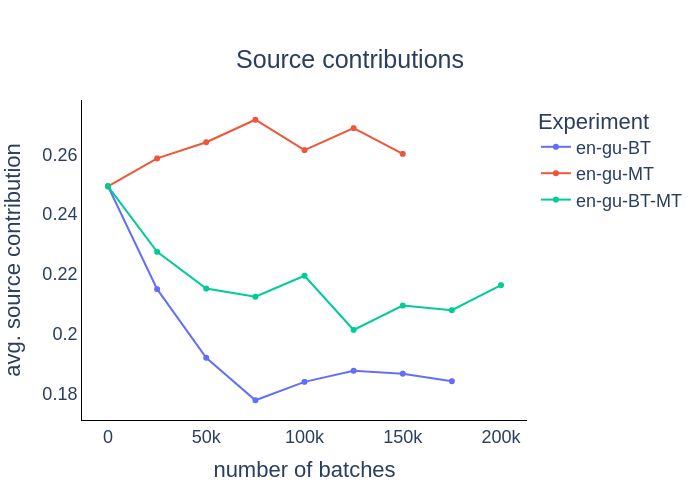}
    \includegraphics[width=.33\textwidth]{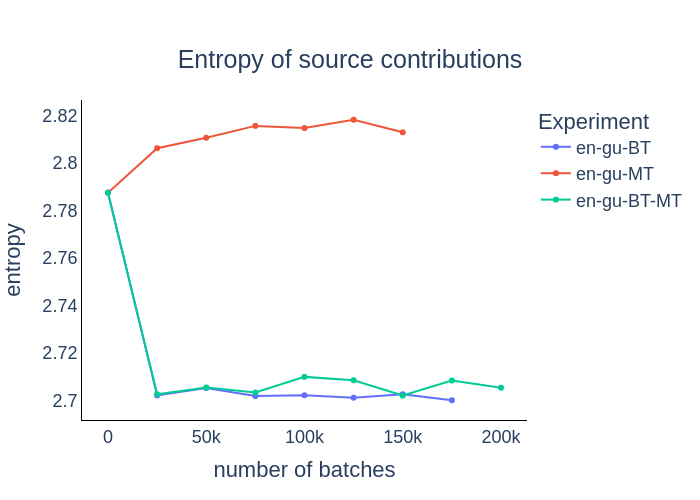}
    \includegraphics[width=.33\textwidth]{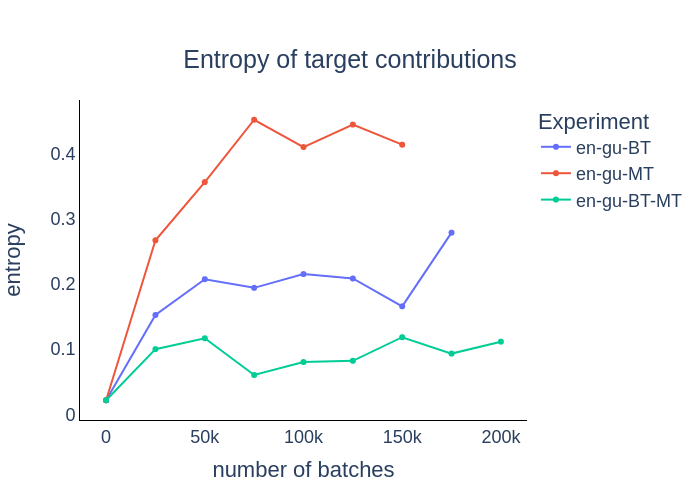}
\end{multicols}
\vspace{-7mm}
\caption{\label{fig:contrib3}En--Gu Average Source Contribution, Entropy of Source Contributions and Entropy of Target Contributions during training.}
\vspace{-4mm}
\begin{multicols}{3}
    \includegraphics[width=.33\textwidth]{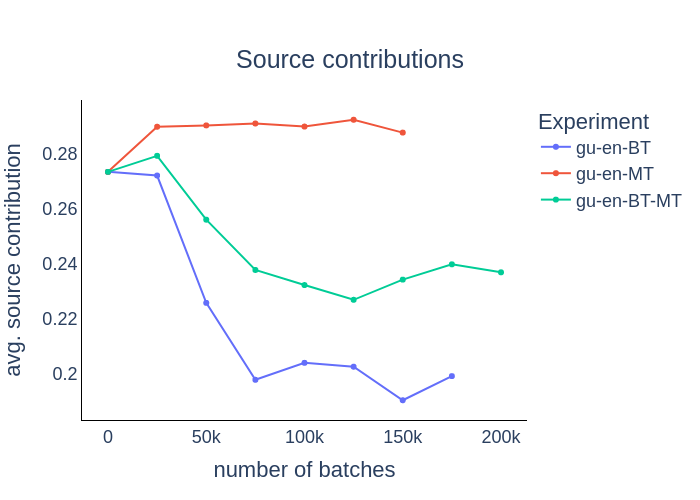}
    \includegraphics[width=.33\textwidth]{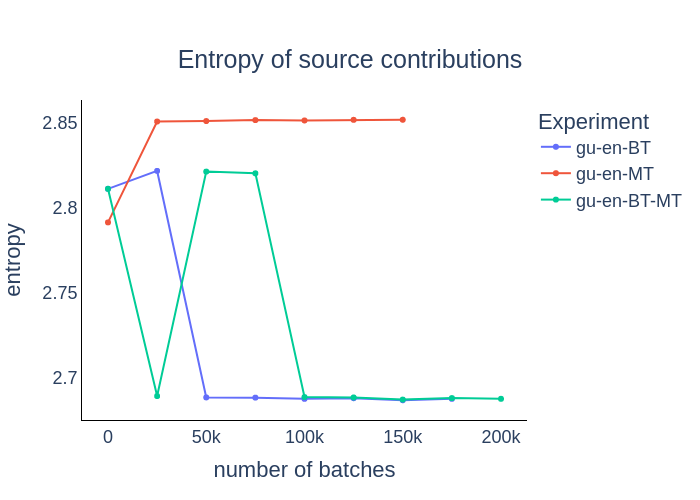}
    \includegraphics[width=.33\textwidth]{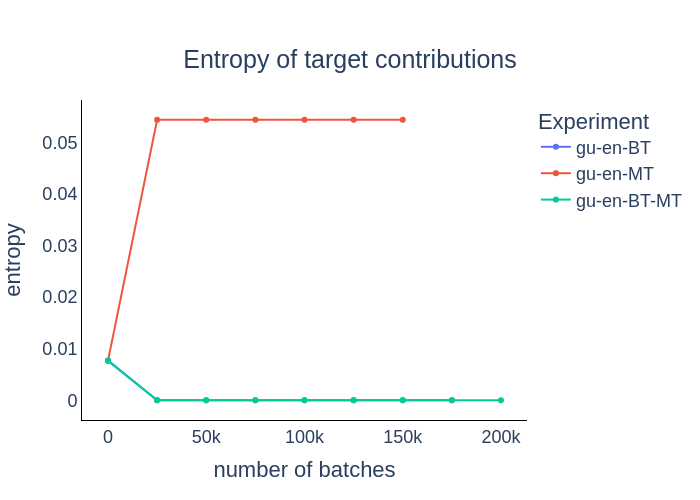}
\end{multicols}
\vspace{-4mm}
\begin{multicols}{3}
    \includegraphics[width=.33\textwidth]{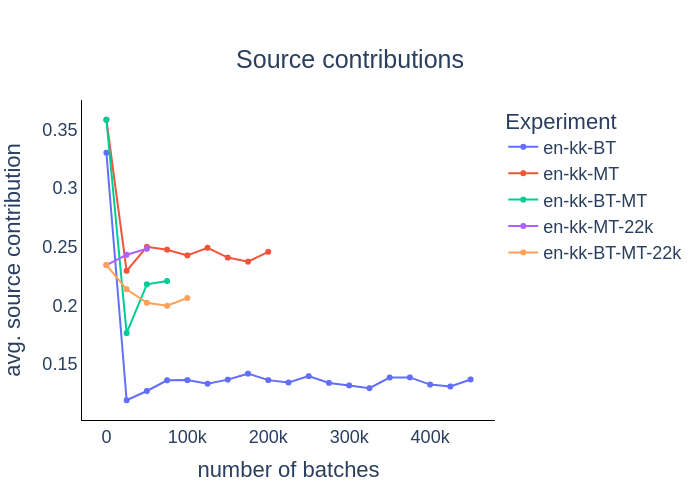}
    \includegraphics[width=.33\textwidth]{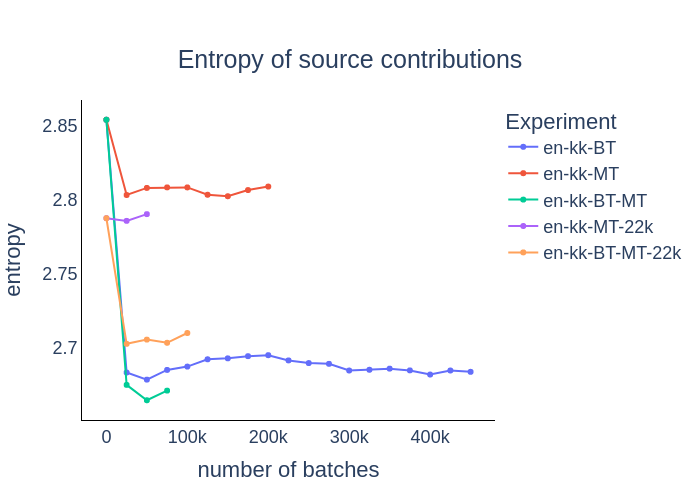}
    \includegraphics[width=.33\textwidth]{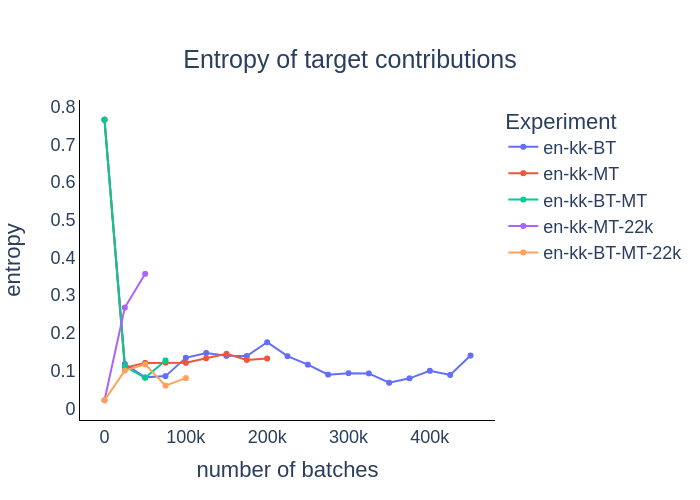}
\end{multicols}
\vspace{-4mm}
\caption{\label{fig:contrib4}En--Kk Average Source Contribution, Entropy of Source Contributions and Entropy of Target Contributions during training.}
\begin{multicols}{3}
    \includegraphics[width=.33\textwidth]{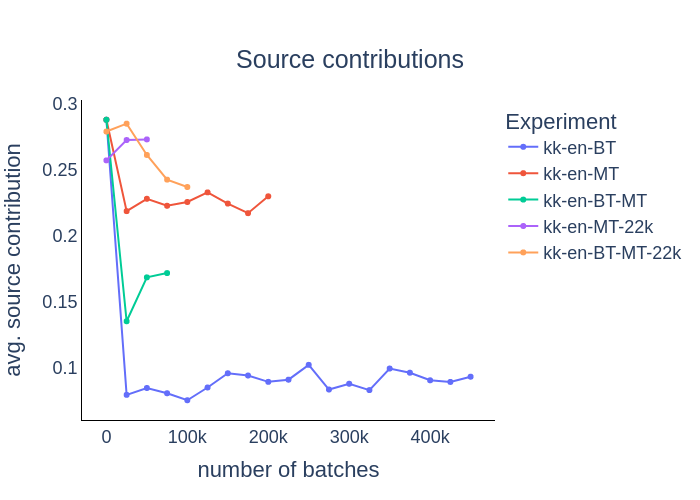}
    \includegraphics[width=.33\textwidth]{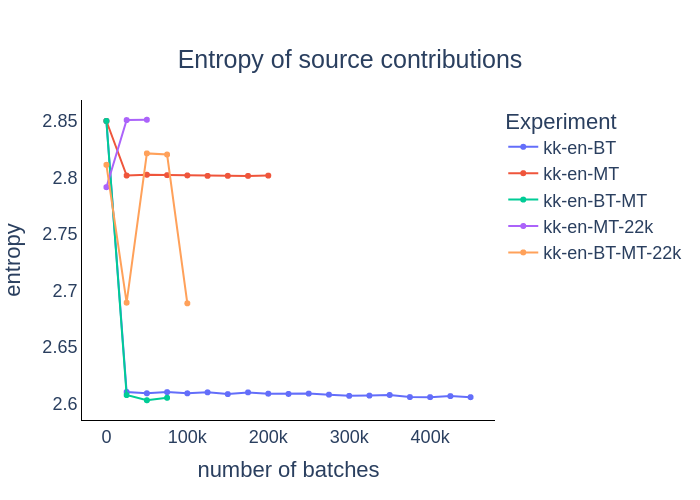}
    \includegraphics[width=.33\textwidth]{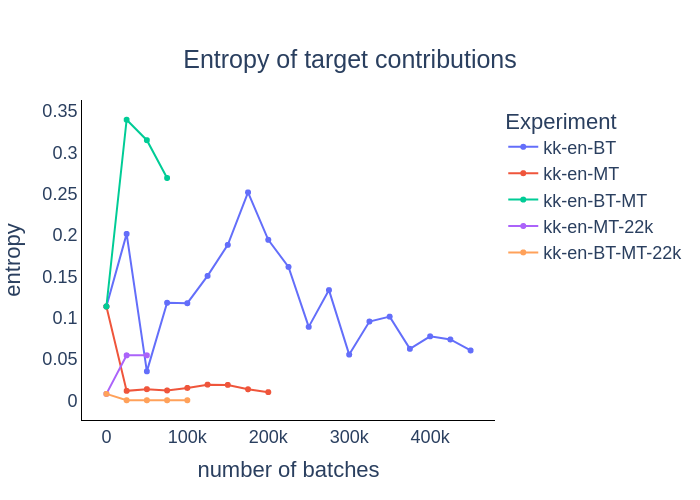}
\end{multicols}
\vspace{-7mm}
\caption{\label{fig:contrib5}Kk--En Average Source Contribution, Entropy of Source Contributions and Entropy of Target Contributions during training.}
\end{figure*}

\vspace{-1mm}
Our observations from average source contribution, entropy of source contributions and entropy of target contributions during training confirm the findings of \citet{voita2020analyzing,voita2021language}. Changes in sentence contributions are not necessarily monotonic   
to the result, can help distinguish different training stages, and identify the balance between source and target sequences' relevance to the result (Fig. \ref{fig:contrib1}, \ref{fig:contrib2}, \ref{fig:contrib3}, \ref{fig:contrib4}).

For En--Fr and Fr--En,  En--Kk and Kk--En NMT models 
(Fig. \ref{fig:contrib1}, \ref{fig:contrib2}) average source sentence contributions drop at the very beginning of training, while contributions are lowest in both directions in MT, and slightly higher in BT, BT-MT experiments; using only parallel (natural) data during training, average source contributions are lower \citep{voita2020analyzing} and the model relies more on the target prefix for sequence generation, while BT boosts the influence of source sentence to the result. 
Average contributions are mostly stable or slightly decrease as training progresses, and the source sentence becomes less important in sequence generation.
For models trained with less data, contributions and relevance of the source sentence tokens to the generated sentence is high, due to the lack of substantial supervision. 

Entropy of source contributions is high for MT-only experiments, 
contributions are more focused, and the model is more confident in choosing the important source tokens, while in BT-only and BT-MT experiments it requires broader source context for target sequence generation, and entropy of contributions is high, for both evaluation directions. In MT-setups, training converges faster.

Studying the entropy of the target contributions, in both En--Fr and Fr--En directions, target entropy is more focused during the first part of training. We then notice either a small (BT, MT-22k, BT-MT-22k) or a larger (MT, BT-MT) increase, which gradually evens out as the model converges. Experiments with a small amount of training data, and/or BT have significantly lower entropy contributions than MT-only, with BT contributing to the model having higher confidence in choosing the target tokens generated. On the contrary, in Fr--En, combined experiments seem to have the highest, hence less focused target contributions; 

Contributions' patterns are not similar for En--Gu and Gu--En models (Fig. \ref{fig:contrib4}, \ref{fig:contrib5}).
Average source contributions in MT experiments 
are higher than those with BT, implying that using parallel data in training forces the model to rely on source tokens more heavily. Average source contributions are lowest in BT-only experiment and target sentence reliance for generation is highest.

Patterns in entropy of source contributions resemble those in En--Fr, Fr--En experiments. Entropy is low in MT-only; training with parallel data increases model confidence in selecting the important source tokens for target generation, while entropy in BT, BT-MT experiments is similarly high. We notice an increase in entropy of target contributions and high values in MT-only experiments in both directions, which validates our hypothesis that source contributions are more focused in these cases 
while the entropy in BT experiments is lower.
Looking for differences between evaluation directions, En--Gu contributions in MT- and BT-MT are similar to those in the En--Fr low resource experiments, in contrast to training in the other direction.

We conclude that back-translation (BT) boosts the influence of source sentences, particularly in low-resource settings, while also highlighting that sentence contributions are not necessarily monotonic and indicate different training stages.

In Tables \ref{table:examples_fr}, \ref{table:examples_gu} in the Appendix we show a few example sentences and their translations, at the beginning and end of training of each model. Examples of sentences and their perturbations are given in Table \ref{table:examples_perturbed}.

\section{Conclusions}
\label{conclusions}

We conduct an extensive analysis of Supervised and/or Unsupervised NMT models’ behavior for French, Gujarati and Kazakh NMT, to and from English, and examine the output in terms of quality, word order, semantic similarity and reliance on source and reference sentences. Our results highlight the importance of supervision for output quality, yet outline the superiority of UNMT in generating sentences highly aligned to references and in preserving models’ robustness. We hope our work sets the ground for better understanding and improving UNMT and our findings can be utilized to improve real-world UNMT systems.

\section{Limitations}
\label{limitations}
It is a computationally hard task to train
large Neural Machine Translation models from scratch and the complexity of the training process is high, calling for more efficient training solutions, in terms of memory distribution of the model and parallelization. It is strongly recommended to design a more systematic approach to addressing those factors and expand to more languages, in order to achieve further generalization of the method and overcome all current limitations. Moreover, results for low-resource NMT systems may often be poor, or marginally improving state-of-the-art, calling for improvement in NMT methods to boost performance.
\section{Ethical Considerations}
\label{ethics}
The authors of the paper are aware that when training large language models, several issues ought to be taken into account, related to quality, toxicity and bias related to their training process and output \cite{10.1145/3442188.3445922,chowdhery2022palm,brown2020language}.

\clearpage
\bibliography{anthology,custom}
\bibliographystyle{acl_natbib}

\section*{Appendix}
\subsection*{Robustness}
En--Fr and Fr--En NMT models are highly robust in all MT, BT-MT setups (Table \ref{table:bleu_robust_consist}), 
especially when test sets are misspelled. 
On the contrary, En--Gu and Gu--En models are highly robust on BT, BT-MT experiments, 
on test sets perturbed by case-changing; for highly morphological complex languages, BT may help boost model robustness.

A similar behavior is observed for En--Kk, Kk--En. Models are highly robust on case-changing in all unsupervised, supervised and semi-supervised scenarios, and as the amount parallel sentences increases, we see an expected increased robustness to sentences' misspelling; the models become more robust to a high percentage of sentence perturbations with higher training supervision.
In En--Fr and Fr--En NMT models, consistency patterns are similar to those found for Robustness: models are highly consistent in MT, BT-MT experiments, primarily when test sentences are misspelled, with their consistency increasing by the amount of parallel train data. BT-only training does not seem to help.
Model consistency patterns for En--Gu, Gu--En follow those of Robustness, with BT outperforming other methods.


\setlength\rotFPtop{150pt}
\begin{sidewaystable*}[htbp]
\captionsetup{font=scriptsize}
\small
\centering
\vspace{2mm}
\begin{tabular}{l l lll|lll|lll|lll|lll|lll|lll}
\toprule
&&& \textbf{En--Fr} & &  & \textbf{Fr--En} & & & \textbf{En--Gu} &&& \textbf{Gu--En} &&&\textbf{En--Kk}&&&\textbf{Kk--En}& \\\hline
 &  & BLEU & R & C & BLEU & R & C & BLEU & R & C & BLEU & R & C& BLEU & R & C& BLEU & R & C \\\hline
BT  &&&&&&&&&&&&&&&&\\
\textbullet \hspace{0.1cm} original &&\textbf{22.6}&-&-&\textbf{21.78}&-&-&0.31&-&-&0.36&-&-&0.7&-&-&1.0&-&-\\
\textbullet \hspace{0.1cm} misspelling &&14.77&0.65&\textbf{17.82}&16.86&\textbf{0.77}&\textbf{16.52}&\textbf{2.49
}&0.03&\textbf{1.59}&\textbf{3.27}&0.08&\textbf{1.33}&\textbf{1.5}&0.14&\textbf{0.8}&0.7&0.7&\textbf{1.2}\\
\textbullet \hspace{0.1cm} case-changing &&14.87&\textbf{0.66}&13.34&14.56&0.66&11.3&1.22&\textbf{0.93}&0.45&0.91&\textbf{0.52}&0.65&1.2&\textbf{0.71}&0.62&\textbf{1.8}&\textbf{0.8}&0.84\\\hline \hline
22k  &&&&&&&&&&&&&&&& \\\hline
MT  &&&&&&&&&&&&&&&&\\
\textbullet \hspace{0.1cm} original &&\textbf{31.12}&-&-&\textbf{30.63}&-&-&\textbf{2.51}&-&-&\textbf{0.77}&-&-&\textbf{2.4}&-&-&\textbf{2.6}&-&-\\
\textbullet \hspace{0.1cm} misspelling &&30.22&\textbf{0.97}&\textbf{26.03}&30.4&\textbf{0.99}&\textbf{31.33}&0.05&\textbf{0.01}&0&0.23&0.29&0.3&1.4&0.58&1.1&1.2&0.46&\textbf{1.3}\\
\textbullet \hspace{0.1cm} case-changing &&20.01&0.64&22.13&23.34&0.76&24.93&0&0&0&0.33&\textbf{0.42}&\textbf{0.47}&1.9&\textbf{0.79}&1.3&2.2&\textbf{0.84}&1.9\\\hline
BT+AE+MT &&&&&&&&&&&&&&&&\\
\textbullet \hspace{0.1cm} original &&\textbf{34.42}&-&-&\textbf{33.87}&-&-&\textbf{1.08}&-&-&2.19&-&-&2.8&-&-&2.9&-&-\\
\textbullet \hspace{0.1cm} misspelling &&31.79&\textbf{0.92}&\textbf{28.18}&32.62&\textbf{0.96}&\textbf{33.44}&0.72&0.66&\textbf{0.79}&\textbf{3}&0.36&\textbf{2.37}&\textbf{3.2}&0.14&1.7&3.0&0.02&\textbf{3.1}\\
\textbullet \hspace{0.1cm} case-changing &&23.06&0.66&22.52&27&0.79&24.22&0.9&\textbf{0.83}&0.6&2.12&\textbf{0.96}&1.88&2.5&\textbf{0.89}&\textbf{1.83}&\textbf{3.5}&\textbf{0.2}&2.7\\\hline \hline
132k &&&&&&&&&&&&&&&&\\\hline
MT  &&&&&&&&&&&&&&&&\\
\textbullet \hspace{0.1cm} original &&\textbf{37.8}&-&-&\textbf{35.6}&-&-&-&-&-&-&-&-&5.2&-&-&8.0&-&-\\
\textbullet \hspace{0.1cm} misspelling &&35.2&\textbf{0.93}&\textbf{30.2}&33.8&\textbf{0.94}&\textbf{35.3}&-&-&-&-&-&-&4.9&0.94&14.6&7.8&\textbf{0.97}&7.5\\
\textbullet \hspace{0.1cm} case-changing &&24.01&0.63&23.1&25.2&0.7&25.4&-&-&-&-&-&&5.0&\textbf{0.96}&4.7&7.7&0.96&\textbf{7.6}\\\hline
BT+AE+MT  &&&&&&&&&&&&&&&&\\
\textbullet \hspace{0.1cm} original &&\textbf{38.6}&-&-&\textbf{38.4}&-&-&-&-&-&-&-&-&5.8&-&-&8.9&-&-\\
\textbullet \hspace{0.1cm} misspelling &&36.9&0.95&32.5&36.2&\textbf{0.94}&\textbf{34.7}&-&-&-&-&-&-&\textbf{4.7}&\textbf{0.81}&5.4&8.2&\textbf{0.92}&8\\
\textbullet \hspace{0.1cm} case-changing &&25.1&0.68&24.5&26.0&0.67&26.1&-&-&-&-&-&-&4.2&0.72&4.0&8&0.89&\textbf{8.4}\\\hline \hline
23m &&&&&&&&&&&&&&&&\\\hline
MT  &&&&&&&&&&&&&&&&\\
\textbullet \hspace{0.1cm} original &&\textbf{41.84}&-&-&\textbf{41.41}&-&-&-&-&-&-&-&-&-&-&-&-&-&-\\
\textbullet \hspace{0.1cm} misspelling &&40.16&\textbf{0.96}&\textbf{35.36}&40.63&\textbf{0.98}&\textbf{41.5}&-&-&-&-&-&-&-&-&-&-&-&-\\
\textbullet \hspace{0.1cm} case-changing &&27.26&0.65&29.68&30.08&0.72&3\textbf{0.79}&-&-&-&-&-&-&-&-&-\\\hline
BT+AE+MT  &&&&&&&&&&&&&&&&\\
\textbullet \hspace{0.1cm} original &&\textbf{42.63}&-&-&\textbf{42.37}&-&-&-&-&-&-&-&-&-&-&-&-&-&-\\
\textbullet \hspace{0.1cm} misspelling &&39.71&\textbf{0.93}&\textbf{35.04}&40.72&\textbf{0.96}&\textbf{41.49}&-&-&-&-&-&-&-&-&-&-&-&-\\
\textbullet \hspace{0.1cm} case-changing &&28.12&0.65&27.31&33.58&0.79&30.33&-&-&-&-&-&-&-&-&-&-&-&-\\
\bottomrule
\end{tabular}
\caption{BLEU scores, Robustness (R) and Consistency (C) values for Unsupervised (BT), Supervised (MT), and Unsupervised + Supervised (BT+AE+MT) NMT experiments, for the converged model for En--Fr, Fr--En, En--Gu, Gu--En and En--Kk, Kk--En.  Test and validation sets are from WMT19 for Gujarati and Kazakh, and newstest 2013-14 for French, and are perturbed following the method suggested in \citep{niu2020evaluating} for misspelling and case-changing.}
\label{table:bleu_robust_consist}
\end{sidewaystable*}

\subsection*{Semantic Similarity}
\begin{figure*}[htb]
\captionsetup{font=scriptsize}
\begin{multicols}{4}
 \includegraphics[width=.33\textwidth]{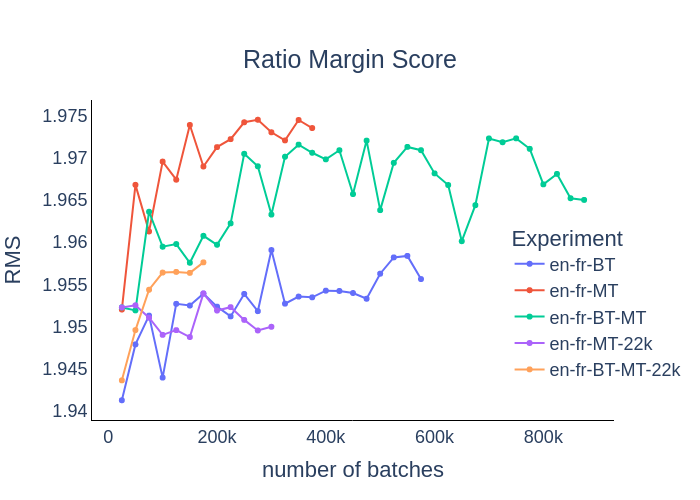}
 \includegraphics[width=.33\textwidth]{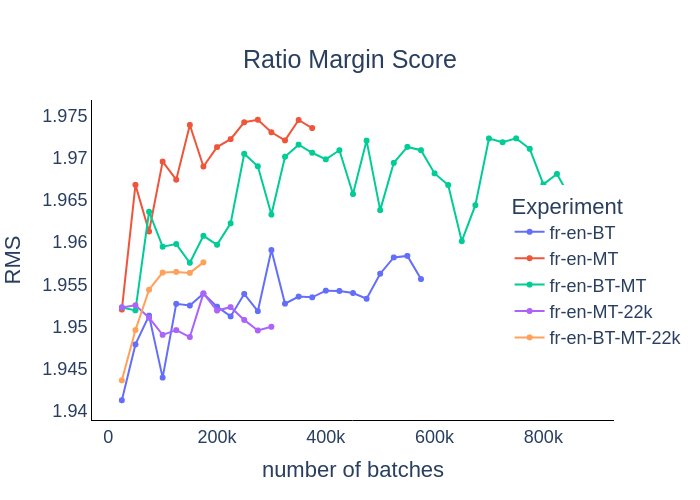}
 \includegraphics[width=.33\textwidth]{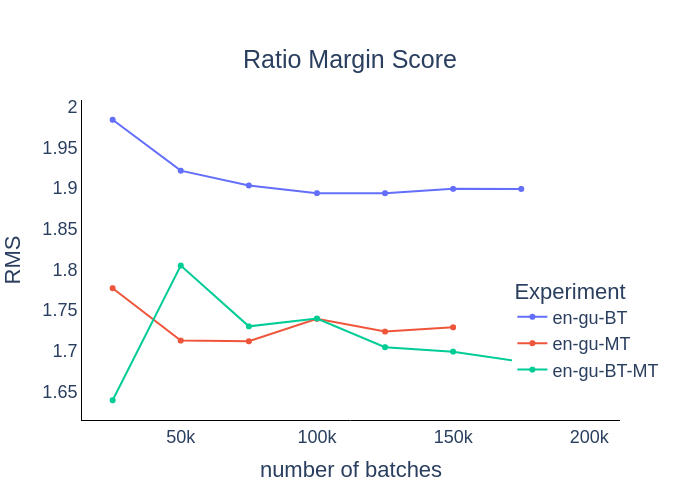}
\includegraphics[width=.33\textwidth]{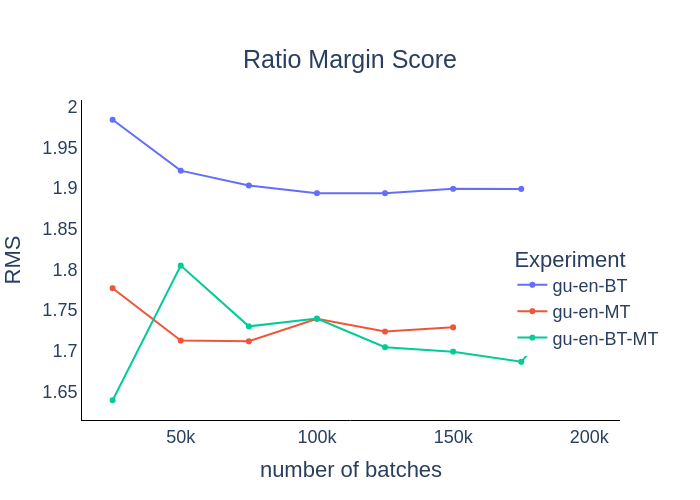}
\includegraphics[width=.33\textwidth]{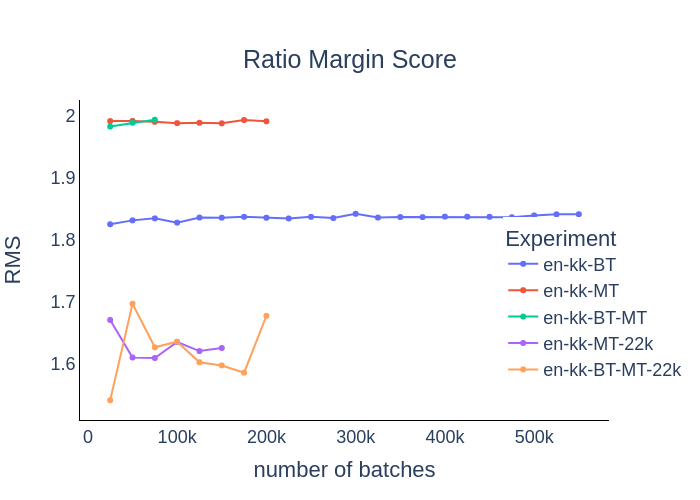}
\includegraphics[width=.33\textwidth]{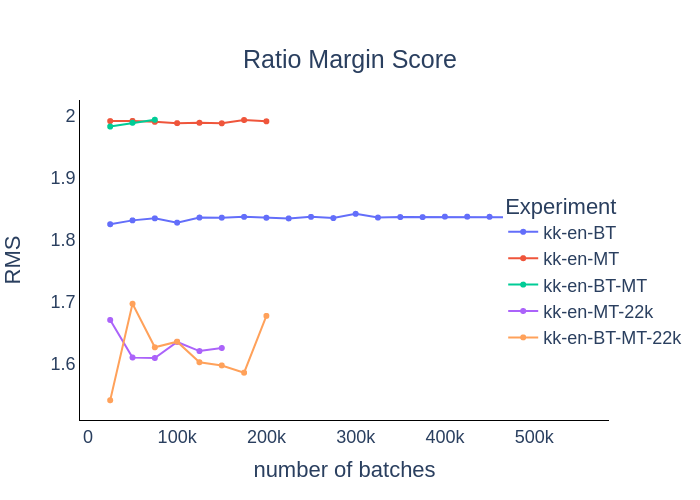}
\end{multicols}
\vspace{-6mm}
\caption{RMSS between references and generated translations for En--Fr, Fr--En, En--Gu, Gu--En, En--Kk, Kk--En during training.}
\label{fig:rms_over_batches}
\vspace{-4mm}
\begin{multicols}{4}
\includegraphics[width=.33\textwidth]{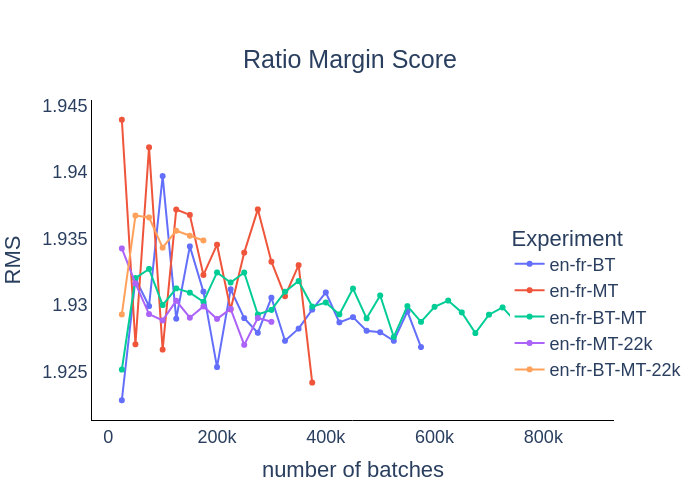}
\includegraphics[width=.33\textwidth]{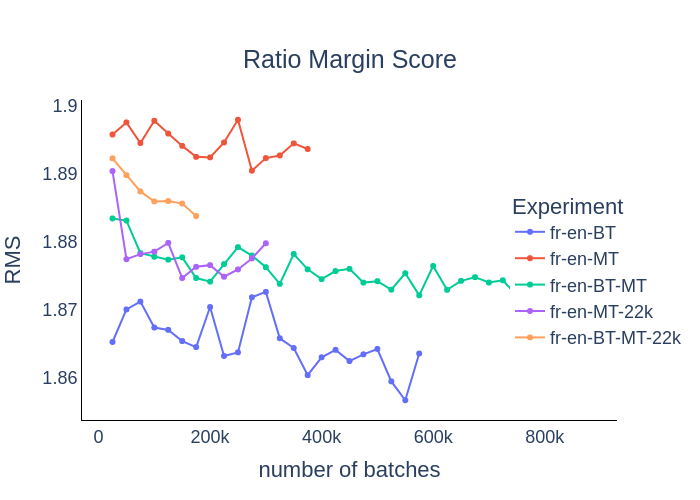}
\includegraphics[width=.33\textwidth]{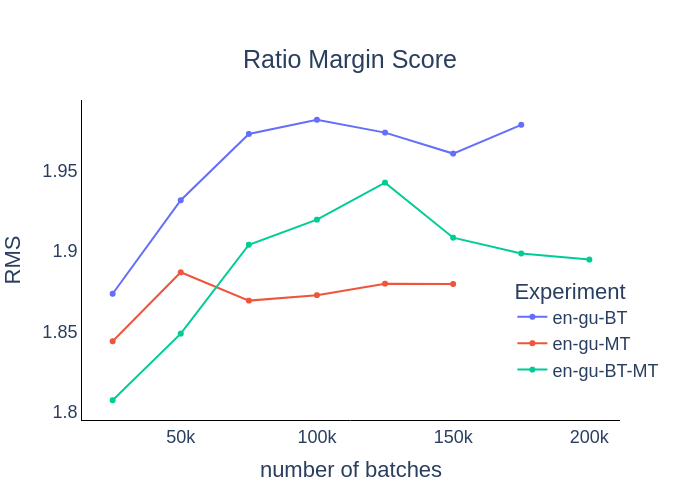}
\includegraphics[width=.33\textwidth]{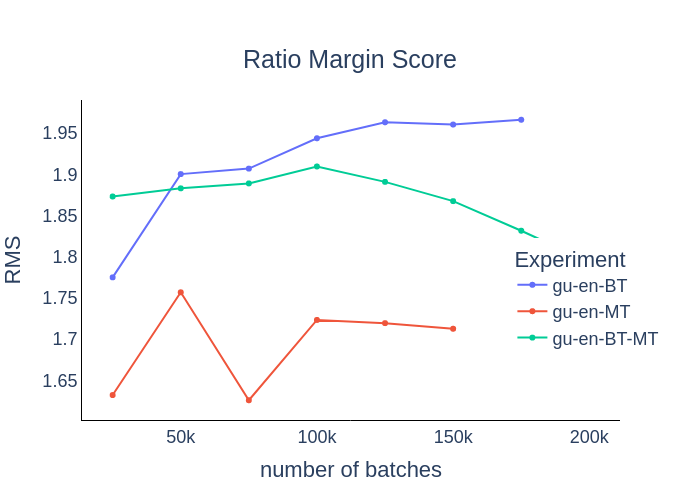}
\includegraphics[width=.33\textwidth]{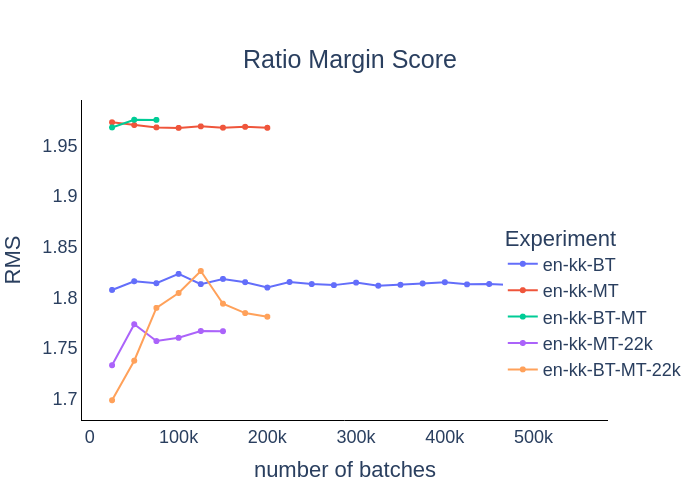}
\includegraphics[width=.33\textwidth]{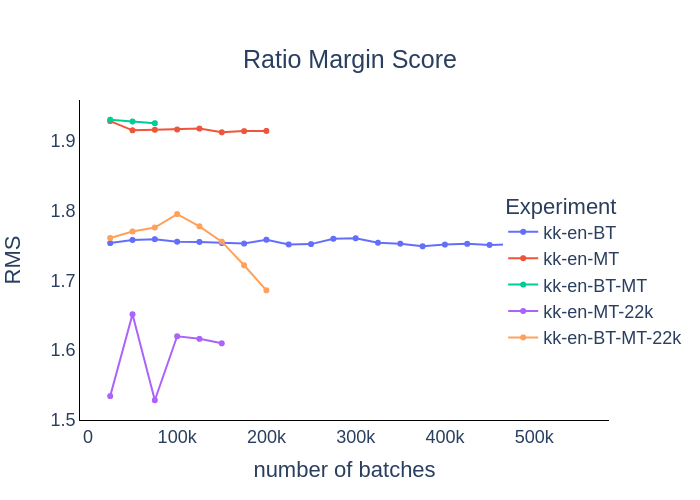}
\end{multicols}
\vspace{-6mm}
\caption{RMSS between source sentences and generated translations for En--Fr, Fr--En, En--Gu, Gu--En, En--Kk, Kk--En during training.}
\label{fig:rms_diff_over_batches}
\end{figure*}

MT-only and BT-MT experiments show high RMSS values in En--Fr, Fr--En between translations and references (Fig. \ref{fig:rms_over_batches}), which have a higher semantic similarity 
than in BT-only or in reduced dataset experiments, 
On the contrary, in En--Gu and Gu--En, translations from MT models are less similar to references, and most similar in BT-only experiments, for which RMSS is highest.
Source sentences show high semantic similarity to translations in MT-only experiments, followed by reduced-data model training results, outperforming BT-only or BT-MT models, in En--Fr and Fr--En; in the first direction, RMSS is very similar across models, while in the latter, behavior of the model in different setups is significantly more distinct.  For En--Gu and Gu--En, BT-only experiments show highest semantic similarity between source sentences and translations (Fig. \ref{fig:rms_diff_over_batches}). 

\clearpage
\begin{table*}[ht]
    \centering
    \small
    \begin{adjustbox}{max width=\textwidth}
        \begin{tabular}{lll}
            \toprule
            \multirow{2}{*}{Original Sentence} & (\textbf{fr}) la certification mondiale de la polio éradication & (\textbf{fr}) dans notre région, la démocratie est une valeur centrale \\
            & (\textbf{en}) global certification of polio eradication & (\textbf{en}) in our region, democracy is a fundamental value \\ 
            \midrule\midrule
            \multirow{2}{*}{\textbf{Model }} & \multicolumn{2}{c}{\textbf{English Translation}} \\ 
            \midrule\midrule
            \multirow{2}{*}{MT-22k} & \textbf{en - FC}: world world dication dication conference & \textbf{en - FC}: in our region, democracy is a core value \\
            & \textbf{en - LC}: world world prohibition of poliomyelitis & \textbf{en - LC}: democracy in our region is a central value \\
            \midrule
            \multirow{2}{*}{BT-MT-22K} & \textbf{en - FC}: global eradication of geromyelite & \textbf{en - FC}: in our region, democracy is a central value \\
            & \textbf{en - LC}: global eradication of poliomyelitis & \textbf{en - LC}: in our region, democracy is a central value \\
            \midrule
            \multirow{2}{*}{BT} & \textbf{en - FC}: the global eradication of poliomyelite & \textbf{en - FC}: In our region, democratie is a central value \\
            & \textbf{en - LC}: the global eradication of poliomyelite & \textbf{en - LC}: In our region, democratie is a central value \\
            \midrule
            \multirow{2}{*}{MT} & \textbf{en - FC}: the global eradication of poliomyelitis & \textbf{en - FC}: in our region, democracy is a central value \\
            & \textbf{en - LC}: global eradication of polio & \textbf{en - LC}: democracy our region has a central value \\
            \midrule
            \multirow{2}{*}{BT-MT} &  \textbf{en - FC}: global eradication of poliomyelitis &  \textbf{en - FC}: in our region, democracy is a central value \\
            & \textbf{en - LC}: global eradication of poliomyelitis & \textbf{en - LC}: democracy our region is a central value \\
            \bottomrule
        \end{tabular}
    \end{adjustbox}
    \vspace*{-0.5em}
    \caption{Two examples of a French sentence, their English ground-truth translation, and their English translations with each model's first and last checkpoint (\textbf{en} - FC, \textbf{en} - LC respectively).}
    \vspace{-2mm}
    \label{table:examples_fr}
    \vspace{-0.5em}
\end{table*}

\begin{table*}[ht]
\centering
\begin{adjustbox}{max width=\textwidth}
\begin{tabular}{lll}
\toprule
\multirow{2}{*}{Original sentence} & \textbf{gu} & \includegraphics[scale=0.5]{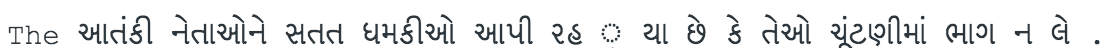} \\
& \textbf{en} & The militants are constantly intimidating the politician not to participate in the election. \\
\midrule\midrule
\multirow{2}{*}{BT} & \textbf{en} - FC & \includegraphics[scale=0.5]{gu_sentences/gu_bt_hyp0.png} \\
& \textbf{en} - LC & \includegraphics[scale=0.5]{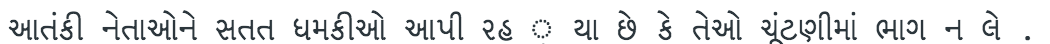} \\
\midrule
\multirow{2}{*}{MT} & \textbf{en} - FC & \includegraphics[scale=0.3]{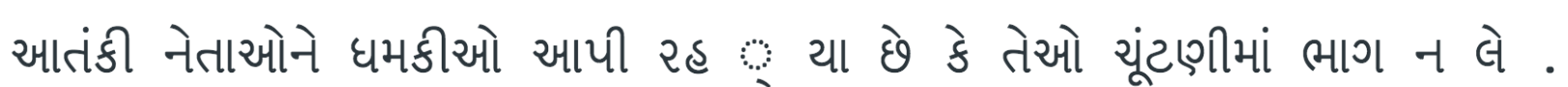} \\
& \textbf{en} - LC & \includegraphics[scale=0.3]{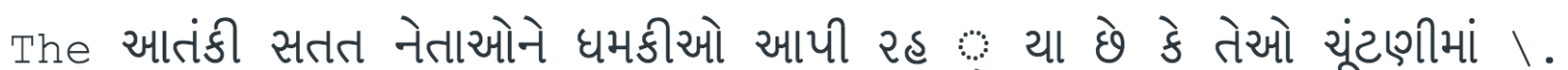} \\
\midrule
\multirow{2}{*}{BT-MT} & \textbf{en} - FC & \includegraphics[scale=0.5]{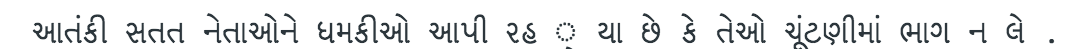} \\
& \textbf{en} - LC & \includegraphics[scale=0.5]{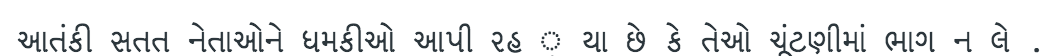} \\
\midrule\midrule
\multirow{2}{*}{Original sentence} & \textbf{gu} & \includegraphics[scale=0.5]{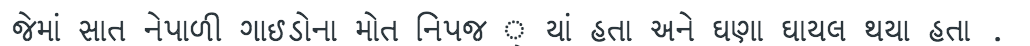} \\
& \textbf{en} & In which seven Nepalese guides were found dead and many were injured. \\
\midrule\midrule
\multirow{2}{*}{BT} & \textbf{en} - FC & \includegraphics[scale=0.5]{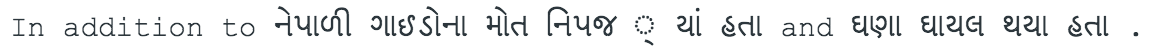} \\
& \textbf{en} - LC & \includegraphics[scale=0.5]{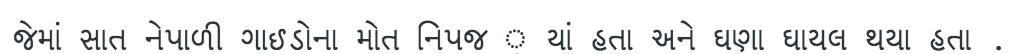} \\
\midrule
\multirow{2}{*}{MT} & \textbf{en} - FC & \includegraphics[scale=0.3]{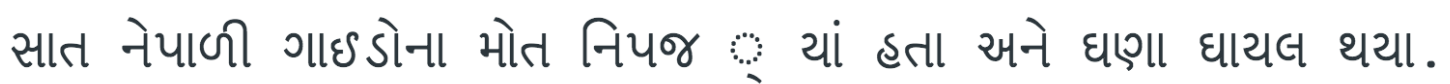} \\
& \textbf{en} - LC & \includegraphics[scale=0.3]{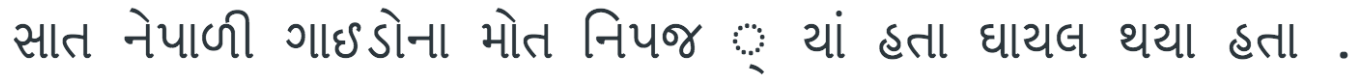} \\
\midrule
\multirow{2}{*}{BT-MT} & \textbf{en} - FC & \includegraphics[scale=0.5]{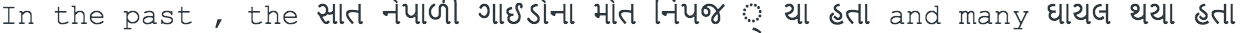} \\
& \textbf{en} - LC & \includegraphics[scale=0.5]{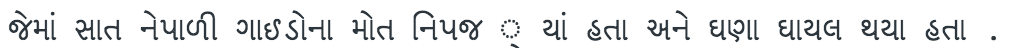} \\
\midrule\midrule
\bottomrule
\end{tabular}
\end{adjustbox}
\vspace*{-0.5em}
\caption{Two examples of a Gujarati sentence, their English ground-truth translation, and their English translations with each model's first and last checkpoint (\textbf{en} - FC, \textbf{en} - LC respectively)}
\vspace{-2mm}
\label{table:examples_gu}
\vspace{-0.5em}
\end{table*}

\clearpage
\setlength\rotFPtop{150pt}
\begin{sidewaystable}[htbp]
\captionsetup{font=scriptsize}
\tiny
\begin{tabular}{lll}
\toprule
\multirow{2}{*}{} & & \\
& & \\

\multirow{3}{*}{\textbf{Original (En)}} & 
despite the measures taken by developing countries to enhance the dissemination of information and strengthen regulations and surveillance of financial markets, & \\
& they remained extremely vulnerable to economic crises and international macroeconomic cycles. & \\
\multirow{2}{*}{\textbf{Misspelling}} & 
despute the measures taken by developing countries to enhance the dissemination of information and strengthen regulations and surveillance of financial markets, & \\
& they remained extremely vulnerable to economic crises and international macroeconomic cycles. & \\
\multirow{2}{*}{\textbf{Case-changing}} & 
Despite The Measures Taken By Developing Countries To Enhance The Dissemination Of Information And Strengthen Regulations And Surveillance Of Financial Markets, & \\
& They Remained Extremely Vulnerable To Economic Crises And International Macroeconomic Cycles. & \\
\midrule\midrule

\multirow{3}{*}{\textbf{Original (Fr)}} & 
il a ete souligne que, par souci de coherence, le libelle de cet alinea devrait etre aligne sur celui du paragraphe 4 concernant & \\
& la constitution d'une garantie dans le contexte des mesures provisoires inter partes, si ce n'est que les mots "peut faire obligation" pourraient etre remplaces par les mots "fait obligation". & \\
\multirow{2}{*}{\textbf{Misspelling}} & 
l a ete souligne que, par souci de coherence, le libelle de cet alinea devrait etre aligne sur celui du paragraphe 4 concernant & \\
& la constitution d'une garantie dans le contexte des mesures provisoires inter partes, si ce n'est que les mots ``peut faire obligation`` pourraient etre remplaces par les mots ``fait obligation``. & \\
\multirow{2}{*}{\textbf{Case-changing}} & 
IL A ETE SOULIGNE QUE, PAR SOUCI DE COHERENCE, LE LIBELLE DE CET ALINEA DEVRAIT ETRE ALIGNE SUR CELUI DU PARAGRAPHE 4 CONCERNANT & \\
& LA CONSTITUTION D'UNE GARANTIE DANS LE CONTEXTE DES MESURES PROVISOIRES INTER PARTES, SI CE N'EST QUE LES MOTS "PEUT FAIRE OBLIGATION" & \\ 
& POURRAIENT ETRE REMPLACES PAR LES MOTS "FAIT OBLIGATION". & \\
\midrule\midrule
\multirow{2}{*}{\textbf{Original (Gu)}} & 
\includegraphics[scale=0.5]{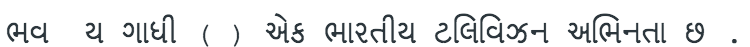} & \\
\multirow{2}{*}{\textbf{Misspelling}} & 
\includegraphics[scale=0.5]{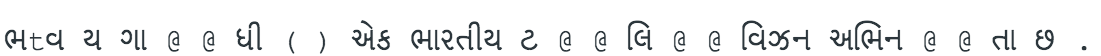} & \\
\multirow{2}{*}{\textbf{Case-changing}} & 
\includegraphics[scale=0.5]{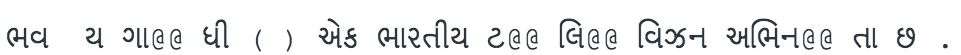} & \\
\midrule
\bottomrule
\end{tabular}
\vspace*{-0.5em}
\caption{Examples of sentences in our test dataset, and their perturbed versions after misspelling and case-changing}
\vspace{-2mm}
\label{table:examples_perturbed}
\vspace{-0.5em}
\end{sidewaystable}

\end{document}